\documentclass[10pt,twocolumn,letterpaper]{article}
\PassOptionsToPackage{table}{xcolor}
\usepackage[pagenumbers]{cvpr}

\usepackage{makecell}
\usepackage{microtype}
\usepackage{mathtools}

\newcommand{\ind}{\mathbf{1}}
\newcommand{\sg}{\operatorname{sg}}

\newcommand{\NA}{\textemdash}
\definecolor{oursred}{RGB}{252,232,232}
\newcommand{\cmark}{\ensuremath{\checkmark}}
\newcommand{\xmark}{\ensuremath{\times}}
\newcommand{\oursrow}{\rowcolor{oursred}}

\definecolor{cvprblue}{rgb}{0.21,0.49,0.74}
\usepackage[pagebackref,breaklinks,colorlinks,allcolors=cvprblue]{hyperref}

\usepackage{amsmath}
\usepackage{booktabs}
\usepackage{multirow}
\usepackage{graphicx}
\usepackage{algorithm}
\usepackage{algpseudocode}
\usepackage{xcolor}
\usepackage{capt-of}
\usepackage{cancel}
\usepackage{tabularx}
\usepackage{array}
\usepackage[table]{xcolor}
\usepackage{placeins}
\def\confName{CVPR}
\def\confYear{2027}
\newcommand{\method}{{\upshape\scshape SPACE-SET}}

\title{\method: Staged Pixel-Space Adaptation of Conditioning\\and EO Generation for Low-Step SAR-to-EO Translation}

\newcommand{\corrauthor}{\textsuperscript{\ensuremath{\dagger}}}
\newcommand{\authoremail}[1]{\href{mailto:#1}{\textcolor{black}{\small\texttt{#1}}}}
\author{%
  \begin{tabular*}{\textwidth}{@{}*{3}{>{\centering\arraybackslash}p{.333333\textwidth}}@{}}
    Hyunseo Lee & Jeonghyeok Do\corrauthor & Munchurl Kim\corrauthor \\
    KAIST & KAIST & KAIST \\
    \authoremail{heart2002101@kaist.ac.kr} &
    \authoremail{ehwjdgur0913@kaist.ac.kr} &
    \authoremail{mkimee@kaist.ac.kr}
  \end{tabular*}%
}
\hypersetup{
  pdftitle={SPACE-SET: Staged Pixel-Space Adaptation of Conditioning and EO Generation for Low-Step SAR-to-EO Translation},
  pdfauthor={Hyunseo Lee, Jeonghyeok Do, Munchurl Kim}
}
\begin{document}
\typeout{=== TEXTWIDTH: \the\textwidth ===}
\typeout{=== COLUMNWIDTH: \the\columnwidth ===}
\raggedbottom

\makeatletter
\renewcommand{\@maketitle}{%
    \noindent\begin{minipage}{\textwidth}
        \centering
        {\Large\bfseries \@title\par}
        \vspace{13pt}
        {\large\setlength{\tabcolsep}{0pt}\@author\par}
        \vspace{5pt}
        \url{https://kaist-viclab.github.io/SPACE-SET_site}\par
    \end{minipage}\par
    \vspace{9pt}
    \noindent
    \begin{minipage}{\textwidth}
        \centering
        \includegraphics[width=\textwidth]
        {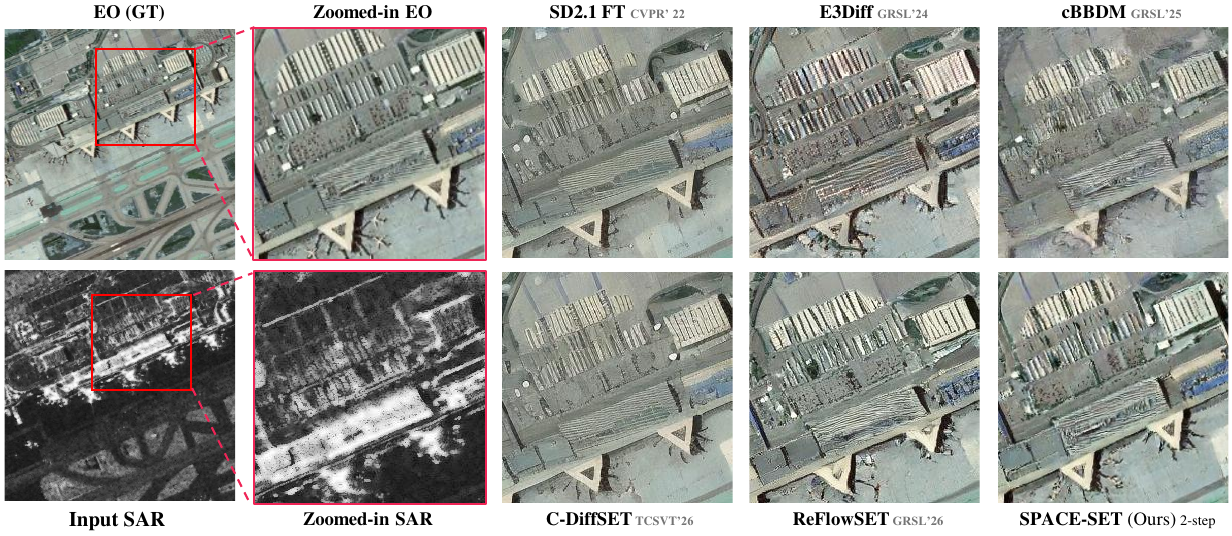}
        \captionof{figure}{
            \textbf{High-fidelity SAR-to-EO translation with only two sampling steps.}
            Qualitative comparison with representative diffusion- and flow-based methods.
            Enlarged regions (red boxes) highlight the improved building structures and fine roof details generated by \method{}.
        }
        \label{fig:teaser}
    \end{minipage}\par
    \vspace{10pt}
}
\makeatother

\maketitle
\begingroup
\renewcommand{\thefootnote}{\fnsymbol{footnote}}
\footnotetext[2]{Corresponding authors.}
\endgroup

\begin{abstract}
Latent generative models have shown promise in SAR-to-Electro-Optical Translation (SET), but their reliance on variational autoencoders (VAEs) can limit fine-grained fidelity and low-step inference efficiency.
Recent advances in pixel-space diffusion, particularly direct clean-image prediction from noisy pixels, have renewed interest in VAE-free generation and motivated adapting pretrained latent generators.
We introduce \textbf{\method{}} (\textbf{S}taged \textbf{P}ixel-Space \textbf{A}daptation of \textbf{C}onditioning and \textbf{E}O Generation for SAR-to-EO Translation), a staged framework that converts a pretrained latent generator into a VAE-free pixel-space model.
To avoid disrupting pretrained representations through simultaneous modality conversion, EO Pixel Transfer (EPT) first adapts the EO generation pathway to pixel space while preserving the pretrained SAR conditioning pathway. SAR Pixel Adaptation (SPA) then converts the SAR conditioning pathway and jointly adapts the generator using pixel-wise uncertainty flow matching (PUFM) and Context-Preserving Distillation (CPD) from the EPT teacher.
Both phases share a three-stage curriculum progressing from flow learning and representation alignment to perceptual and adversarial supervision.
Low-Step Flow Adaptation (LFA) then separately fine-tunes one- and two-step generators from the SPA model under teacher guidance.
Extensive experiments show that our two-step generator leads on four of five quality metrics on QXS-SAROPT, while our one-step generator maintains competitive translation quality with \textbf{76.07$\times$} faster inference than a state-of-the-art SET baseline.
\end{abstract}
\section{Introduction}
\label{sec:intro}

Optical remote sensing imagery provides intuitive visual appearance and rich spectral information for interpreting land cover and built structures, but cloud cover and illumination limit its acquisition. In contrast, synthetic aperture radar (SAR) enables day-and-night, all-weather observations, yet speckle noise and complex scattering make direct interpretation difficult, particularly for non-experts~\cite{WANG2026105009, LI202114, rs18010073, Samadzadegan01022025}. SAR-to-electro-optical translation (SET) generates visually interpretable EO-like images from SAR, potentially enabling pretrained optical-domain models for downstream tasks such as semantic segmentation, object detection, and vision-language understanding~\cite{11134789, carion2025sam3segmentconcepts, liu2023grounding, kuckreja2024geochat, liu2024remoteclip, 11609477}. However, SAR and EO measure different physical properties, while misregistration and acquisition-time differences further complicate their correspondence. These challenges call for SET models that generate realistic EO images, preserve structures captured by SAR, and efficiently process large SAR datasets over wide areas.

Early SET methods employed GAN-based conditional or unpaired image-to-image translation~\cite{zhao2022comparative,YANG2022108208,9750140,he2025dogan}. Although GANs enable fast, typically single-pass inference, they often struggle with fine textures and spatial structures in complex scenes. Diffusion-based methods improved realism and structural fidelity through iterative denoising, including conditional DDPMs and diffusion-bridge formulations~\cite{10330015,10744588,10623721,kim2025conditional}. One-step diffusion-based SET models have also been explored to reduce sampling costs~\cite{qin2024efficient}. Latent diffusion and flow-based approaches further reduce per-step computation by generating in compact VAE-encoded representations while leveraging pretrained generative priors~\cite{rombach2021highresolution,do2026c,do2026reflowset}. However, VAE compression can discard fine-scale details, while a fixed decoder limits the image details recoverable through image-level supervision. Moreover, fixed VAE encoding and decoding costs become increasingly significant in low-step generation, limiting the speedup achievable by reducing sampling steps alone.

To overcome these limitations, recent studies have revisited pixel-space generative models that synthesize images directly without relying on pretrained VAEs~\cite{Hoogeboom_2025_CVPR,Ma_2026_CVPR,li2026back,chen2026dip,yu2026pixeldit}. Without a fixed VAE decoder, pixel-space generation enables direct image-level supervision for fine-grained detail preservation while eliminating VAE overhead, particularly benefiting low-step inference. However, training high-dimensional pixel-space generators from scratch remains computationally demanding. To address this cost, recent latent-to-pixel transfer methods convert pretrained latent generators into VAE-free pixel-space models while reusing their learned generative priors~\cite{chen2026l2p,wang2026eliminating}. However, in conditional SET, simultaneously converting the SAR conditioning and EO generation pathways may disrupt their learned interactions, limiting the effective reuse of pretrained representations.

\begin{figure}[t]
    \centering
    \includegraphics[width=\columnwidth]{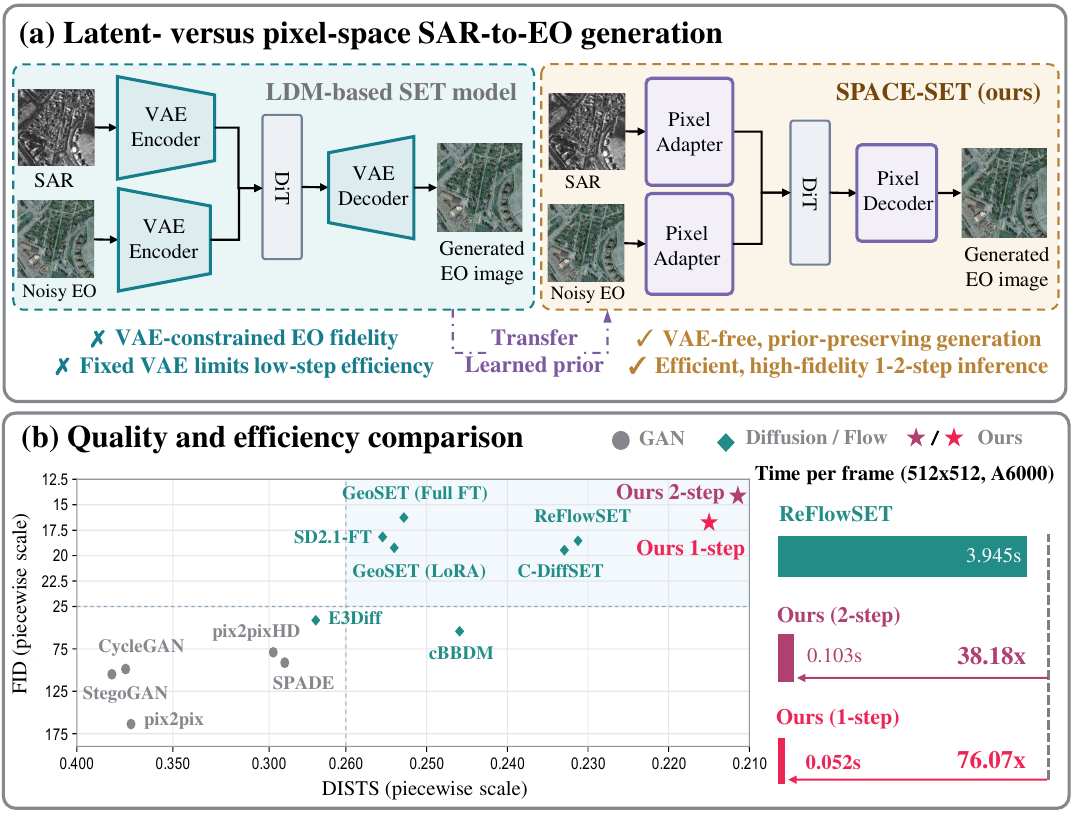}

    \caption{\textbf{Progressive pixel-space conversion for efficient SET.}
    \textbf{(a)} Conventional latent-space generation vs. our VAE-free pixel-space framework.
    \textbf{(b)} Translation quality (FID and DISTS on QXS-SAROPT) vs. inference latency (SAR2Opt).}
    \label{fig:latent_pixel}
    \vspace{-10px}
\end{figure}

Motivated by these advances, we propose \method{}, a staged pixel-space adaptation framework for efficient low-step SET. EO Pixel Transfer (EPT) first converts EO generation to pixel space while retaining pretrained SAR conditioning, followed by SAR Pixel Adaptation (SPA), which converts the remaining SAR pathway to enable fully VAE-free generation (Fig.~\ref{fig:latent_pixel} (a)). Low-Step Flow Adaptation (LFA) then enables efficient one- and two-step generation through teacher-guided adaptation. Fig.~\ref{fig:teaser} compares our results with state-of-the-art (SOTA) methods, highlighting improved scene structures and fine details, while Fig.~\ref{fig:latent_pixel} (b) illustrates the quality--efficiency trade-off using FID and DISTS on QXS-SAROPT and inference latency on SAR2Opt. Experiments on QXS-SAROPT~\cite{huang2021qxs} and SAR2Opt~\cite{zhao2022comparative} validate our staged adaptation strategy.

Our contributions are summarized as follows:
\begin{itemize}
    \item We propose \method{}, a \textbf{staged pixel-space adaptation framework comprising EPT, SPA, and LFA}, which \textit{progressively} converts a pretrained \textit{latent} SAR-to-EO generator into a VAE-free \textit{pixel-space} model and enables efficient one- and two-step generation.

    \item We design a \textbf{remote-sensing-oriented three-stage curriculum} whose objectives are shared across \textbf{EPT, SPA, and LFA}, progressively introducing representation alignment, perceptual learning, and adversarial refinement during pixel-space conversion.

    \item \method{} achieves \textbf{high-fidelity SET with only two sampling steps} on QXS-SAROPT and SAR2Opt, while its one-step generator provides a \textbf{$76.07\times$ inference speedup} over ReFlowSET with competitive quality.
\end{itemize}

\section{Related Work}
\label{sec:related}

\subsection{General Image-to-Image Translation}

Image-to-image translation learns cross-domain mappings under paired or unpaired supervision. pix2pix~\cite{isola2017image} combines a U-Net generator, PatchGAN discriminator, and reconstruction loss for paired translation, while CycleGAN~\cite{zhu2017unpaired} enables unpaired translation through adversarial learning and cycle consistency. For high-resolution synthesis, pix2pixHD~\cite{wang2018pix2pixHD} employs coarse-to-fine generation, multi-scale discrimination, and feature matching, while SPADE~\cite{8953676} preserves semantic layouts through spatially adaptive normalization. StegoGAN~\cite{wu2024stegogan} separates matchable and unmatchable features using steganography to prevent spurious structures in non-bijective translation. More recently, latent diffusion models~\cite{rombach2021highresolution} leverage pretrained VAE representations to provide transferable generative priors.

\subsection{SAR-to-EO Image Translation}

SET faces additional challenges due to different SAR--EO imaging properties and imperfect spatial correspondence. Early GAN-based methods address these challenges through specialized architectures. ICGAN~\cite{YANG2022108208} combines parallel feature fusion, multi-scale discrimination, and color-aware supervision, while FG-GAN~\cite{9750140} employs asymmetric generation and cross-modal normalization for unpaired translation. DOGAN~\cite{he2025dogan} further leverages pretrained DINO optical priors through multi-scale feature alignment.

Diffusion-based methods improve generation through SAR-conditioned denoising~\cite{10330015}, spatial-frequency refinement~\cite{10744588}, and color-aware Brownian bridges~\cite{10623721}. cBBDM~\cite{kim2025conditional} extends conditional Brownian bridge diffusion to very-high-resolution SET, while E3Diff~\cite{qin2024efficient} combines SAR spatial priors and adversarial refinement for one-step generation. C-DiffSET~\cite{do2026c} adapts pretrained latent diffusion with confidence-guided supervision, while OSCAR~\cite{10.1007/978-3-032-31438-3_27} integrates optical semantic distillation and uncertainty-aware guidance. ReFlowSET~\cite{do2026reflowset} analyzes SAR--EO VAE reconstruction fidelity to select a suitable latent codec and trains a compact conditional flow-matching model with representation alignment. Despite improving latent reconstruction quality, ReFlowSET remains dependent on the VAE, motivating our progressive conversion toward VAE-free generation.

\subsection{Pixel-Space Generation and Latent-to-Pixel Transfer}

Pixel-space generation directly models RGB images without relying on pretrained VAEs, avoiding their reconstruction bottlenecks. JiT~\cite{li2026back} demonstrates effective pixel-space generation through clean-image prediction using large-patch Transformers. However, training pixel-space generators from scratch remains computationally demanding. To reuse pretrained generative priors, L2P~\cite{chen2026l2p} transfers latent diffusion models to pixel space by replacing VAE interfaces with large-patch tokenization while retaining pretrained backbone layers. GenDR-Pix~\cite{wang2026eliminating} progressively removes the VAE encoder and decoder through pixel-(un)shuffle and multi-stage adversarial distillation for image super-resolution.

While these methods demonstrate effective VAE-free generation, extending latent-to-pixel transfer to SET requires preserving learned interactions between SAR conditioning and EO generation under imperfect cross-modal correspondence. To address this, our \method{} sequentially converts the EO and SAR pathways with remote-sensing-oriented supervision, preserving pretrained cross-modal knowledge before adapting the resulting pixel-space generator for efficient one- and two-step inference.

\section{Preliminaries}
\label{sec:preliminaries}

\begin{figure*}[t]
    \centering
    \includegraphics[width=\textwidth]{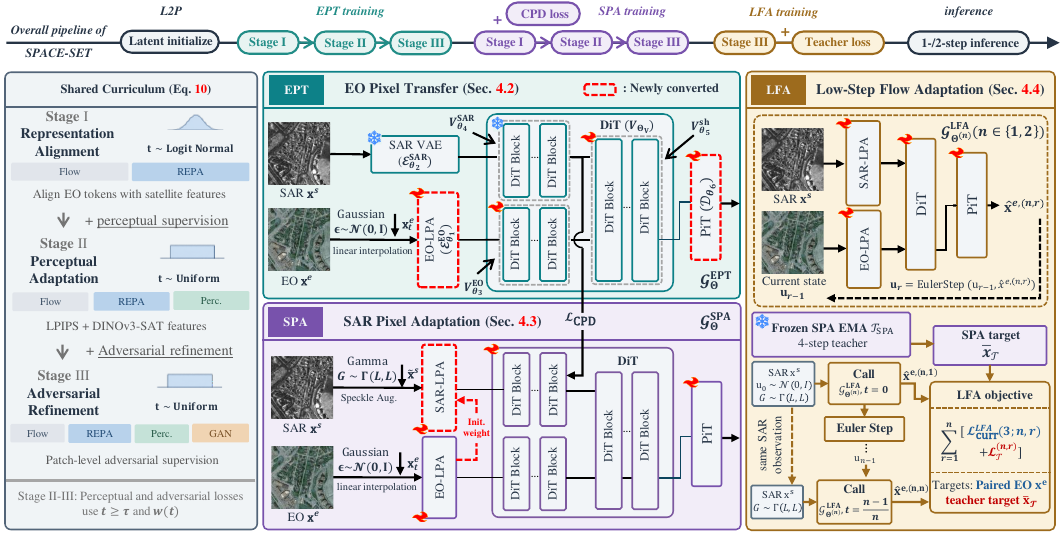}
  
        \caption{\textbf{Sequential conversion and low-step adaptation in \method{}.}
        A shared three-stage curriculum supports \textbf{EPT},
        which converts EO generation while retaining latent SAR
        conditioning, and \textbf{SPA}, which converts SAR
        conditioning through joint adaptation, PUFM, and
        Context-Preserving Distillation.
        \textbf{LFA} independently adapts one- and two-step
        generators with teacher guidance, using a connected
        Euler trajectory for two-step training.}
        
    \label{fig:method_overview}
    
\end{figure*}

\subsection{Dual-Stream DiT for SAR-to-EO Translation}
\label{sec:prelim_reflowset}

We adopt ReFlowSET~\cite{do2026reflowset} as our pretrained backbone, which performs SET through conditional latent flow matching with a dual-stream DiT. Given a SAR observation $\mathbf x^s$ and paired EO image $\mathbf x^e$, modality-specific VAE encoders $\mathcal E_{\theta_2}^{\mathrm{SAR}}$ and $\mathcal E_{\theta_1}^{\mathrm{EO}}$ produce their latent representations $\mathbf z^s$ and $\mathbf z^e$, respectively. The EO and SAR streams, $V_{\theta_3}^{\mathrm{EO}}$ and $V_{\theta_4}^{\mathrm{SAR}}$, interact through joint attention; their concatenated output tokens pass through shared DiT blocks $V_{\theta_5}^{\mathrm{sh}}$. The EO VAE decoder $\mathcal D_{\theta_6}$ decodes the predicted EO latent. Schematically, the generator and its modality-specific pathways are
\begin{equation}
\begin{gathered}
\mathcal G_\Theta
=\mathcal D_{\theta_6}\circ
\underbrace{
V_{\theta_5}^{\mathrm{sh}}\circ
\overbrace{
(V_{\theta_3}^{\mathrm{EO}}\Vert V_{\theta_4}^{\mathrm{SAR}})
}^{\text{joint attention}}
}_{V_{\Theta_V}}
\circ(\mathcal E_{\theta_1}^{\mathrm{EO}},\mathcal E_{\theta_2}^{\mathrm{SAR}}),\\[5pt]
\mathcal H_{\Theta_E}^{\mathrm{EO}}
\triangleq(\mathcal E_{\theta_1}^{\mathrm{EO}},V_{\theta_3}^{\mathrm{EO}}),\quad
\mathcal H_{\Theta_S}^{\mathrm{SAR}}
\triangleq(\mathcal E_{\theta_2}^{\mathrm{SAR}},V_{\theta_4}^{\mathrm{SAR}}).
\end{gathered}
\label{eq:generator_composition}
\end{equation}
Here, $V_{\Theta_V}$ is the full DiT backbone, $\mathcal H_{\Theta_E}^{\mathrm{EO}}$ and $\mathcal H_{\Theta_S}^{\mathrm{SAR}}$ are the EO generation and SAR conditioning pathways, respectively, and $\Vert$ denotes concatenation between streams. The functional input interfaces $\mathcal E_{\theta_1}^{\mathrm{EO}}$, $\mathcal E_{\theta_2}^{\mathrm{SAR}}$, and output decoder $\mathcal D_{\theta_6}$ change implementations during pixel-space adaptation.

\subsection{Clean-Image Prediction with Flow Matching}
\label{sec:prelim_fm}

Flow matching~\cite{lipman2022flow} typically learns a noise-to-data velocity field, whereas JiT~\cite{li2026back} directly predicts the clean image from an intermediate noisy state. We enable JiT-style pixel-space training by progressively replacing both modality-specific VAE input interfaces with bottleneck-structured Lightweight Pixel Adapters (LPAs), EO-LPA and SAR-LPA.

For Gaussian noise $\boldsymbol\epsilon\sim\mathcal N(\mathbf0,\mathbf I)$, the linear noise-to-EO interpolation path is
\begin{equation}
\mathbf x_t^e=(1-t)\boldsymbol\epsilon+t\mathbf x^e,
\qquad t\in[0,1].
\label{eq:prelim_flow_path}
\end{equation}
Given the noisy EO input $\mathbf x_t^e$ processed by EO-LPA and SAR observation $\mathbf x^s$, the generator predicts the clean EO image $\hat{\mathbf x}^e=\mathcal G_\Theta(\mathbf x_t^e,t,\mathbf x^s)$.
The prediction is supervised using the following timestep-dependent flow matching loss:
\begin{equation}
\ell_{\mathrm{FM}}
=(\hat{\mathbf x}^e-\mathbf x^e)^2/d(t)^2,
\quad
\mathcal L_{\mathrm{FM}}=\mathbb E[\ell_{\mathrm{FM}}],
\label{eq:prelim_fm_loss}
\end{equation}
where $d(t)=\max(1-t,0.05)$ prevents the denominator from vanishing near $t=1$.

\section{\method{}}
\label{sec:method}

Fig.~\ref{fig:method_overview} illustrates the \method{} framework, consisting of EO Pixel Transfer (EPT), SAR Pixel Adaptation (SPA), and Low-Step Flow Adaptation (LFA). We denote their generators as $\mathcal G_\Theta^{\mathrm{EPT}}$, $\mathcal G_\Theta^{\mathrm{SPA}}$, and $\mathcal G_{\Theta^{(n)}}^{\mathrm{LFA}}$ ($n\in\{1,2\}$), respectively. Since the EO generation pathway $\mathcal H_{\Theta_E}^{\mathrm{EO}}$ and SAR conditioning pathway $\mathcal H_{\Theta_S}^{\mathrm{SAR}}$ interact through joint attention, simultaneously replacing their VAE-based input interfaces with pixel-space adapters may limit the use of pretrained representations and complicate optimization. To address this, \method{} employs staged pixel-space adaptation: (i) EPT first converts the EO input interface in $\mathcal H_{\Theta_E}^{\mathrm{EO}}$ into pixel space and the EO decoder into pixel space using standard flow matching (FM), while keeping $\mathcal H_{\Theta_S}^{\mathrm{SAR}}$ frozen, yielding a \textit{hybrid pixel-/latent-space generator} (Sec.~\ref{sec:ept}); (ii) SPA then converts the SAR input interface in $\mathcal H_{\Theta_S}^{\mathrm{SAR}}$ into pixel space and jointly adapts the generator using Context-Preserving Distillation (CPD) from the EPT teacher and pixel-wise uncertainty flow matching (PUFM), yielding a \textit{fully pixel-space generator} (Sec.~\ref{sec:spa}); (iii) LFA adapts the converted generator for efficient one- and two-step inference under SPA teacher guidance (Sec.~\ref{sec:lfa}). EPT and SPA follow a shared remote-sensing-oriented three-stage curriculum that begins with flow learning and representation alignment, followed by perceptual and adversarial supervision (Sec.~\ref{sec: pixel-space-curriculum}).

\subsection{Pixel-Space Flow and Shared Curriculum}
\label{sec:architecture}
\label{sec:flow}
\label{sec:transfer}
\label{sec:arch}
\label{sec:curriculum}

\paragraph{Pixel-wise Uncertainty flow supervision.}
\label{sec: pixel-space-curriculum}
To account for pixel-space uncertainty arising from spatial misregistration and acquisition-time differences in paired SAR--EO data, we introduce Pixel-wise Uncertainty Flow Matching (PUFM) adapting $\beta$-NLL~\cite{seitzer2022pitfalls} to pixel-space flow residuals:
\begin{equation}
 \mathcal L_{\mathrm{PUFM}}
 =\mathbb E\!\left[
 \sg[q_v]^{\,\beta}
 \left(\ell_{\mathrm{FM}}/q_v
 +\log(2\pi q_v)\right)\right],
 \label{eq:pufm}
\end{equation}
where a training-only head predicts per-pixel variance $q_v>0$ from the final features of pixel decoder $\mathcal D_{\theta_6}$; $\sg[\cdot]$ stops gradients and $\beta=1$, as detailed in \textit{Supplementary matrial}.

\noindent\textbf{Stage I: representation alignment.}
Representation alignment (REPA)~\cite{yu2024representation}
encourages satellite-aware generator representations
using a frozen satellite foundation encoder
$\mathcal F_\phi$.
A trainable projector $\mathcal P_\omega$ aligns EO patch
features $\mathbf h_e^L$ from the final ($L$-th)
dual-stream block $V_{\theta_3}^{\mathrm{EO}}$
with corresponding last-layer features
$\mathcal F_\phi^{\mathrm{last}}(\mathbf x^e)$
of the clean EO image:
\begin{equation}
\mathcal L_{\mathrm{REPA}}
=1-\mathbb E\!\left[\cos\!\left(
\mathcal P_\omega(\mathbf h_e^L),
\sg[\mathcal F_\phi^{\mathrm{last}}(\mathbf x^e)]
\right)\right].
\label{eq:repa}
\end{equation}
Flow times follow a logit-normal distribution.

\noindent\textbf{Stage II: perceptual adaptation.}
We add output-level supervision using VGG-based
LPIPS~\cite{zhang2018unreasonable} and hierarchical
features from $\mathcal F_\phi$. The satellite perceptual distance averages cosine distances between corresponding patch features over selected hierarchical layers $\mathcal K$ of $\mathcal F_\phi$:
\begin{equation}
\ell_{\mathrm{SAT}}(\mathbf u,\mathbf v)
=\mathbb E_{k\in\mathcal K}\Big[
1-\cos\big(
\mathcal F_\phi^{(k)}(\mathbf u),
\mathcal F_\phi^{(k)}(\mathbf v)
\big)\Big].
\label{eq:sat_perceptual}
\end{equation}
For stability, perceptual supervision is restricted
to $t\geq\tau$, and further weighted by $w(t)=d(t)^{-\gamma}$:
\begin{equation}
\begin{aligned}
\ell_{\mathrm P}(\mathbf u,\mathbf v)
=&\lambda_{\mathrm{LP}}\ell_{\mathrm{LPIPS}}(\mathbf u,\mathbf v)+\lambda_{\mathrm{SAT}}\ell_{\mathrm{SAT}}(\mathbf u,\mathbf v),\\
\mathcal L_{\mathrm{perc}}
&=\mathbb E\!\left[
w(t)\ell_{\mathrm P}(\hat{\mathbf x}^e,\mathbf x^e)
\,\middle|\,t\geq\tau
\right],
\end{aligned}
\label{eq:perceptual_objective}
\end{equation}
where $\tau=0.3$, $\gamma=1$; flow times are sampled uniformly.

\noindent\textbf{Stage III: adversarial refinement.}
We add patch-level adversarial supervision on EO images
using a lightweight two-layer classifier head
$h_{\psi_k}$ for each selected layer $k\in\mathcal K$
of $\mathcal F_\phi$, producing patch logits
$C_{\psi_k}(\mathbf u)
=h_{\psi_k}(\mathcal F_\phi^{(k)}(\mathbf u))$. Applying Stage 2's timestep constraint and weighting gives
\begin{equation}
\begin{aligned}
\ell_{\mathrm{GAN}}(\mathbf u)
&=\sum_{k\in\mathcal K}
\ell_{\mathrm{BCE}}(C_{\psi_k}(\mathbf u),1),\\
\mathcal L_{\mathrm{GAN}}
&=\mathbb E\!\left[
w(t)\ell_{\mathrm{GAN}}(\hat{\mathbf x}^e)
\,\middle|\,t\geq\tau
\right],
\end{aligned}
\label{eq:gan_objective}
\end{equation}
where $\ell_{\mathrm{BCE}}$ denotes binary cross-entropy with logits averaged over patch locations. Flow times remain uniform, and generator EMA is introduced in this stage.

\noindent\textbf{Shared cumulative objective.}
We define the phase-specific flow objective for
$P\in\{\mathrm{EPT},\mathrm{SPA},\mathrm{LFA}\}$ as
\begin{equation}
 \mathcal L_{\mathrm{flow}}^{P}=
 \begin{cases}
  \mathcal L_{\mathrm{FM}}, & P=\mathrm{EPT},\\
  \mathcal L_{\mathrm{PUFM}}, & P\in\{\mathrm{SPA},\mathrm{LFA}\}.
 \end{cases}
 \label{eq:phase_flow}
\end{equation}
The curriculum progressively accumulates supervision:
\begin{equation}
\begin{aligned}
 \mathcal L_{\mathrm{curr}}^{P}(k)
 &={}\mathcal L_{\mathrm{flow}}^{P}
 +\lambda_{\mathrm{REPA}}\mathcal L_{\mathrm{REPA}}\\
 &+\ind[k\geq2]\mathcal L_{\mathrm{perc}}+\ind[k=3]\lambda_{\mathrm{GAN}}\mathcal L_{\mathrm{GAN}},
\end{aligned}
\label{eq:stage_objectives}
\end{equation}
where $k\in\{1,2,3\}$ indexes curriculum stages; EPT and SPA use all three, while LFA uses only the Stage-3 objective with step-specific supervision.

\subsection{EO Pixel Transfer (EPT)}
\label{sec:ept}
\label{sec:conversion}

EPT initializes $V_{\Theta_V}$ from pretrained ReFlowSET~\cite{do2026reflowset}, inheriting its cross-modal representations, and replaces $\mathcal E_{\theta_1}^{\mathrm{EO}}$ and $\mathcal D_{\theta_6}$ with an EO-LPA and PiT-based pixel decoder~\cite{yu2026pixeldit}, respectively. The SAR pathway $\mathcal H_{\Theta_S}^{\mathrm{SAR}}$ remains frozen to retain pretrained latent SAR conditioning, while the remaining parameters $\Theta_{\mathrm{train}}^{\mathrm{EPT}}=\Theta\setminus\Theta_S$ are jointly optimized under the shared three-stage curriculum in Eq.~\ref{eq:stage_objectives}:
\begin{equation}
\mathcal L_{\mathrm{EPT}}(k)
=\mathcal L_{\mathrm{curr}}^{\mathrm{EPT}}(k),
\quad k\in\{1,2,3\}.
\label{eq:ept_objective}
\end{equation}
Importantly, EPT not only transfers the EO generation
pathway to pixel space but also establishes a hybrid
teacher that preserves the pretrained SAR
conditioning interface and its interaction
with the pixel-space EO generation pathway.

\subsection{SAR Pixel Adaptation (SPA)}
\label{sec:spa}

EPT produces a hybrid pixel-/latent-space generator, but its SAR conditioning pathway remains dependent on the VAE encoder. This residual encoding cost is particularly significant for one- and two-step inference. SPA removes this overhead by replacing $\mathcal E_{\theta_2}^{\mathrm{SAR}}$ with a SAR-LPA initialized from the EPT-trained EO-LPA. The EO input interface and decoder retain their pixel-space implementations, while all generator parameters are jointly optimized, $\Theta_{\mathrm{train}}^{\mathrm{SPA}}=\Theta$. We also use Gamma SAR augmentation and timestep-gated SAR detail injection, as detailed in \textit{Supplementary matrial}.

\noindent\textbf{Context-Preserving Distillation.}
We distill SAR-stream features from the frozen EMA EPT teacher $\mathcal T_{\mathrm{EPT}}=\mathcal G_\Theta^{\mathrm{EPT}}$ to the SPA student $\mathcal G_\Theta^{\mathrm{SPA}}$ to preserve cross-modal representations. Let $\mathbf h_S^{\mathrm{ds}}$ and $\mathbf h_T^{\mathrm{ds}}$
denote final dual-stream block features of
the student and teacher SAR pathways, respectively. The student uses speckle-augmented SAR input $\tilde{\mathbf x}^s$, whereas the teacher uses the original input $\mathbf x^s$; both receive the same noisy EO $\mathbf x_t^e$ and timestep $t$. We define CPD as:
\begin{equation}
\mathcal L_{\mathrm{CPD}}
=\mathbb E\!\left[
\|\mathbf h_S^{\mathrm{ds}}-\sg[\mathbf h_T^{\mathrm{ds}}]\|_F^2 / (N_sD_h)
\right],
\label{eq:kd_loss}
\end{equation}
where $N_s$ and $D_h$ are the SAR token count and feature dimension. Since joint attention incorporates EO context into the SAR-stream features, CPD guides SAR-pathway adaptation while preserving the EPT teacher's EO-conditioned representations. The SPA objective is
\begin{equation}
\mathcal L_{\mathrm{SPA}}(k)
=\mathcal L_{\mathrm{curr}}^{\mathrm{SPA}}(k)
+\lambda_{\mathrm{CPD}}\mathcal L_{\mathrm{CPD}},
\quad k\in\{1,2,3\}.
\label{eq:phase_b_objective}
\end{equation}

\subsection{Low-Step Flow Adaptation (LFA)}
\label{sec:lfa}
\label{sec:twostep}

LFA independently fine-tunes one- and two-step students
$\mathcal G_{\Theta^{(n)}}^{\mathrm{LFA}}$ ($n\in\{1,2\}$)
from the fully converted SPA generator, updating all parameters.
Starting from Gaussian noise
$\mathbf u_0=\boldsymbol\epsilon\sim\mathcal N(\mathbf0,\mathbf I)$,
each student makes $n$ sequential clean-image predictions at
$t_{r-1}=(r-1)/n$, updating the pixel-space state through
uniform Euler time steps of size $1/n$.
Thus, one-step LFA directly predicts the clean image at $t=0$,
while two-step LFA follows a connected Euler trajectory at
$t=0$ and $0.5$.
Using the same initial noise $\boldsymbol\epsilon$ as the student,
a frozen SPA EMA teacher
$\mathcal T_{\mathrm{SPA}}=\mathcal G_\Theta^{\mathrm{SPA}}$
provides a four-step Euler endpoint
$\bar{\mathbf x}_{\mathcal T}
=\sg[\operatorname{Euler}_4(
\mathcal T_{\mathrm{SPA}};\boldsymbol\epsilon,\mathbf x^s)]$.
The LFA objective combines the Stage-3 curriculum
$\mathcal L_{\mathrm{curr}}^{(n,r)}
\coloneqq \mathcal L_{\mathrm{curr}}^{\mathrm{LFA}}(3;n,r)$
for the $r$-th prediction of the $n$-step student
with teacher-guided MAE and perceptual supervision:
\begin{equation}
\begin{aligned}
&\mathcal L_{\mathrm{LFA}}^{(n)}
=\sum_{r=1}^{n}\Big[
\mathcal L_{\mathrm{curr}}^{(n,r)}
+\lambda_{\mathcal T,1}
\mathbb E[\ell_1(
\hat{\mathbf x}^{e,(n,r)},\bar{\mathbf x}_{\mathcal T})]\\
&\hspace{-10px}\quad+\lambda_{\mathcal T,P}
\mathbf{1}[t_{r-1}\ge\tau]w(t_{r-1})
\mathbb E\big[
\ell_{\mathrm P}(
\hat{\mathbf x}^{e,(n,r)},\bar{\mathbf x}_{\mathcal T})
\big]\Big],
\end{aligned}
\label{eq:lfa_objective}
\end{equation}
where $\ell_1$ denotes MAE, and we define $\mathcal L_{\mathcal T}^{(n,r)}$ as the sum of the last two
teacher-loss terms. For one-step LFA, we
relax the $t\ge\tau$ threshold at $t=0$ for curriculum
perceptual/adversarial and teacher perceptual losses.

\providecommand{\method}{\textsc{GEO-PACE}}
\providecommand{\ExpPending}{\textemdash}
\providecommand{\cmark}{\ensuremath{\checkmark}}
\providecommand{\xmark}{\ensuremath{\times}}

\definecolor{eptbg}{HTML}{FFF9E8}
\definecolor{spabg}{HTML}{FFF4D9}
\definecolor{lfabg}{HTML}{FCEEF1}
\providecommand{\eptrow}{\rowcolor{eptbg}}
\providecommand{\sparow}{\rowcolor{spabg}}
\providecommand{\lfarow}{\rowcolor{lfabg}}

\definecolor{oursbg}{HTML}{FFF9F2}
\providecommand{\oursrow}{\rowcolor{oursbg}}
\definecolor{intermediatebg}{HTML}{FFF9E8}
\providecommand{\intermediaterow}{\rowcolor{intermediatebg}}

\providecommand{\EPTParams}{\ExpPending}
\providecommand{\EPTLatency}{\ExpPending}
\providecommand{\EPTSpeedup}{\ExpPending}
\providecommand{\SPAParams}{\ExpPending}
\providecommand{\SPALatency}{\ExpPending}
\providecommand{\SPASpeedup}{\ExpPending}
\providecommand{\OneStepParams}{\ExpPending}
\providecommand{\OneStepLatency}{\ExpPending}

\providecommand{\OneStepSpeedup}{$81.27\times$}
\providecommand{\TwoStepParams}{\ExpPending}
\providecommand{\TwoStepLatency}{\ExpPending}
\providecommand{\TwoStepSpeedup}{\ExpPending}

\providecommand{\QualPanel}[1]{%
  \IfFileExists{#1}{%
    \includegraphics[width=0.72in,height=0.72in,keepaspectratio]{#1}%
  }{%
    \fbox{\parbox[c][0.69in][c]{0.67in}{%
      \centering\scriptsize Image\\placeholder}}%
  }%
}

\begin{table*}[!t]
\centering
\caption{\textbf{SAR-to-EO translation quality and inference efficiency on QXS-SAROPT and SAR2Opt.}
Inference efficiency is measured on SAR2Opt, with speedup relative to ReFlowSET, while GAN-based methods are reported for completeness but excluded from efficiency comparisons. $\dagger$ denotes intermediate conversion checkpoints preceding the final LFA models.}
\label{tab:main_results}
\setlength{\tabcolsep}{2.0pt}
\renewcommand{\arraystretch}{1.12}
\resizebox{\textwidth}{!}{%
\begin{tabular}{@{}ll*{10}{r}cccccc@{}}
\toprule
\multirow{2}{*}{\textbf{Method}}
& \multirow{2}{*}{\textbf{Venue}} & \multicolumn{5}{c}{\textbf{QXS-SAROPT}}
& \multicolumn{5}{c}{\textbf{SAR2Opt}}
& \multicolumn{6}{c}{\textbf{Efficiency and model size}}\\
\cmidrule(lr){3-7}\cmidrule(lr){8-12}\cmidrule(l){13-18}
& & FID$\downarrow$ & DISTS$\downarrow$ & LPIPS$\downarrow$
& SSIM$\uparrow$ & PSNR$\uparrow$
& FID$\downarrow$ & DISTS$\downarrow$ & LPIPS$\downarrow$
& SSIM$\uparrow$ & PSNR$\uparrow$
& Steps$\downarrow$ & NFE$\downarrow$ & Params. (M)
& FPS$\uparrow$ & ms$\downarrow$ & Speedup$\uparrow$\\
\midrule

\rowcolor{gray!10}
\multicolumn{18}{@{}l}{\textit{GAN-based methods}}\\

pix2pix~\cite{isola2017image}
& CVPR'17
&174.6&0.373&0.665&0.203&12.33
&261.9&0.347&0.657&0.199&13.39
&1&1&54.5&460.33&2.17&$1816.03\times$\\

CycleGAN~\cite{zhu2017unpaired}
& ICCV'17
&104.4&0.376&0.653&0.262&12.92
&143.5&0.330&0.650&0.178&12.90
&1&1&11.4&84.97&11.77&$335.23\times$\\

pix2pixHD~\cite{wang2018pix2pixHD}
& CVPR'18
&85.7&0.298&0.573&0.358&16.13
&146.3&0.283&0.567&0.268&15.95
&1&1&182.4&90.59&11.04&$357.39\times$\\

SPADE~\cite{8953676}
& CVPR'19
&90.7&0.292&0.599&0.320&14.53
&142.5&0.265&0.597&0.234&14.47
&1&1&92.1&30.65&32.62&$120.93\times$\\

StegoGAN~\cite{wu2024stegogan}
& CVPR'24
&106.8&0.384&0.658&0.254&12.96
&150.1&0.347&0.655&0.158&12.47
&1&1&11.4&85.48&11.70&$337.22\times$\\

\midrule
\rowcolor{gray!10}
\multicolumn{18}{@{}l}{\textit{Diffusion and flow-based methods}}\\

SD2.1 fine-tune~\cite{rombach2021highresolution}
& CVPR'22
&19.1&0.257&0.561&0.348&15.40
&\underline{71.8}&0.211&0.541&0.293&16.24
&50&50&949.6&0.78&1,280.59&$3.08\times$\\

E3Diff~\cite{qin2024efficient}
& GRSL'24
&47.8&0.278&0.530&0.302&16.44
&104.7&0.232&0.529&0.249&16.09
&1&1&192.1&6.34&157.80
&$25.00\times$\\

cBBDM~\cite{kim2025conditional}
& GRSL'25
&50.6&0.246&0.539&\underline{0.372}&16.02
&222.3&0.377&0.571&\textbf{0.361}&\textbf{17.05}
&200&200&292.4&0.22&4,601.45&$0.86\times$\\

C-DiffSET~\cite{do2026c}
& TCSVT'26
&19.9&0.233&0.526&\textbf{0.380}&16.92
&78.1&0.214&0.529&\underline{0.314}&\underline{16.81}
&50&50&949.6&0.78&1,291.12&$3.06\times$\\

ReFlowSET~\cite{do2026reflowset}
& GRSL'26
&19.1&0.231&0.534&0.355&16.09
&\textbf{66.3}&0.185&0.522&0.287&16.06
&50&100&593.4&0.25&3,945.03&$1.00\times$\\

\midrule
\rowcolor{gray!10}
\multicolumn{18}{@{}l}{\textit{SET foundation model}}\\

GeoSET (LoRA)~\cite{do2026geoset}
& arXiv'26
&19.3&0.254&0.564&0.326&14.67
&74.6&0.200&0.535&0.278&15.50
&50&100&--&--&--&--\\

GeoSET (Full FT)~\cite{do2026geoset}
& arXiv'26
&16.9&0.244&0.553&0.332&15.07
&\underline{71.1}&0.196&0.532&0.279&15.66
&50&100&--&--&--&--\\

\midrule
\rowcolor{gray!10}
\multicolumn{18}{@{}l}{{\method{} (\textbf{Ours})}}\\

EPT (4-step)$^\dagger$
& --
&17.9&0.225&0.516&0.360&17.02
&72.3&\textbf{0.179}&\underline{0.506}&0.302&16.25
&4&6&659.4&3.04&329.13&$11.99\times$\\

SPA (4-step)$^\dagger$
& --
&\underline{16.8}&0.221&0.509&0.362&17.00
&74.3&0.182&0.508&0.300&16.20
&4&6&625.3&3.23&309.94&$12.73\times$\\

\addlinespace[2pt]

\lfarow
\textbf{LFA (1-step)}
& --
&17.4&\underline{0.213}&\underline{0.498}&0.365&\underline{17.12}
&80.3&0.182&\textbf{0.495}&0.304&16.61
&1&1&625.3&\textbf{19.28}&\textbf{51.86}
&$\mathbf{76.07}\times$\\

\lfarow
\textbf{LFA (2-step)}
& --
&\textbf{15.5}&\textbf{0.212}&\textbf{0.496}&0.369&\textbf{17.17}
&76.9&\underline{0.181}&\textbf{0.495}&0.307&16.55
&2&2&625.3&\underline{9.68}&\underline{103.33}
&$\underline{38.18}\times$\\

\bottomrule

\end{tabular}}
\end{table*}

\section{Experiments}
\label{sec:experiments}
We evaluate translation quality and inference efficiency
on QXS-SAROPT and SAR2Opt, with controlled ablations
on QXS-SAROPT.

\subsection{Experimental Setup}
\label{sec:exp_setup}

\noindent\textbf{Datasets.}
We evaluate \method{} on two paired SAR-to-EO benchmarks:
QXS-SAROPT~\cite{huang2021qxs} and
SAR2Opt~\cite{zhao2022comparative}.
QXS-SAROPT contains 20,000 image pairs at
$256\times256$ resolution, with 16,001 for training
and 3,999 for evaluation.
SAR2Opt contains 2,077 pairs, split into 1,450
training and 627 evaluation samples.
For SAR2Opt, we use random $512\times512$ crops
for training and center crops for evaluation.

\noindent\textbf{Training protocol.}
All phases use two A6000 GPUs and AdamW~\cite{loshchilov2017decoupled} with a constant learning rate of $5\times10^{-5}$, weight decay $0.01$, betas $(0.9,0.999)$, gradient clipping (max norm $1.0$), and batch sizes of $64$/$16$ for QXS-SAROPT/SAR2Opt, respectively. EPT and SPA each run for 60k updates (20k per stage), sampling flow times from a logit-normal distribution with underlying $\mathcal N(-0.8,0.8^2)$ in Stage 1 and $\mathcal U(0,1)$ in Stages 2--3. LFA independently fine-tunes one- and two-step students for 10k updates each, at $t=0$ and along a connected Euler trajectory at $t\in\{0,0.5\}$, respectively. EMA (decay $0.999$) starts in Stage 3 for EPT/SPA and is maintained throughout LFA. Adversarial training uses a 1:1 generator--discriminator update ratio. SAR conditioning is dropped with probability $0.1$; in SPA/LFA, retained SAR inputs undergo Gamma speckle augmentation with probability $0.5$ ($\kappa=0.5$, $L\sim\mathcal U(\{8,16\})$). We instantiate $\mathcal F_\phi$ with DINOv3-SAT
~\cite{simeoni2025dinov3}, pretrained on SAT-493M,
a collection of 493M RGB satellite images.

\noindent\textbf{Inference configuration.}
EPT and SPA use four Euler steps, with CFG~\cite{ho2021classifierfree} scale 2.5
applied only at the middle two steps, resulting in
six NFEs.
LFA uses independently trained one- and two-step
generators without CFG, requiring one and two NFEs,
respectively.
ReFlowSET uses 50 Euler steps with CFG scale 1.5
at every step, resulting in 100 NFEs.

\noindent\textbf{Evaluation metrics.}
We report FID~\cite{heusel2017gans} for distributional realism, DISTS/LPIPS~\cite{ding2020image, zhang2018unreasonable} for perceptual similarity, and SSIM/PSNR~\cite{wang2004image, fardo2016formal} for pixel fidelity. We also report Euler steps, inference-time parameters, latency, and speedup over ReFlowSET. Steps count sampling updates (Euler updates for flow-based methods), whereas NFEs count actual generator forward evaluations, including conditional and unconditional CFG passes. Latency is measured on SAR2Opt at $512\times512$, batch size one, using a single A6000 GPU.

\subsection{Comparison with Existing Methods}
\label{sec:main_results}

\begin{figure*}[t]
    \centering
    \vspace{-8pt}
    \includegraphics[width=\textwidth]{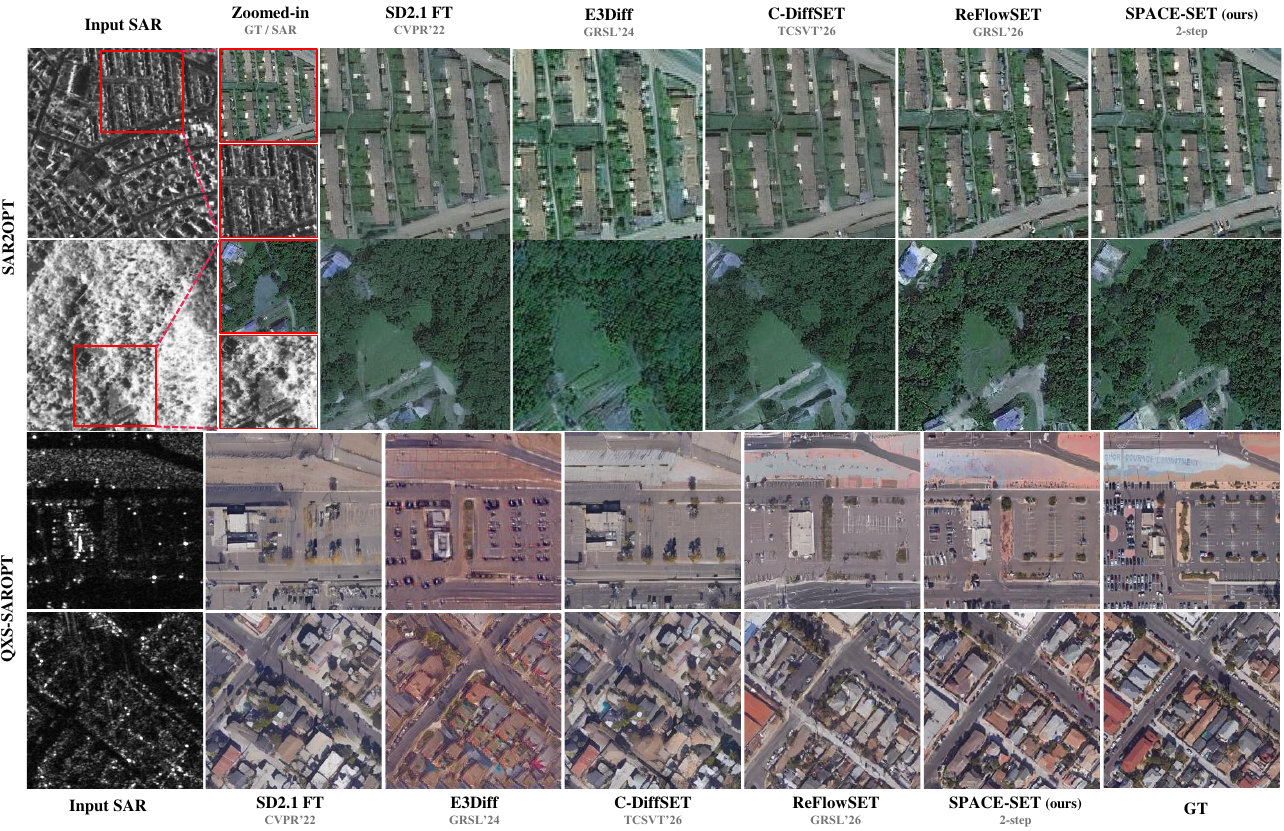}
    \caption{\textbf{Qualitative comparison on SAR2Opt
    (top) and QXS-SAROPT (bottom).}
    SAR2Opt shows zoomed-in comparisons, while QXS-SAROPT
    shows full-image results. \method{} uses two-step inference.}
    \label{fig:qualitative}
    \vspace{-10pt}
\end{figure*}

\noindent\textbf{QXS-SAROPT.}
As shown in Table~\ref{tab:main_results}, our two-step LFA surpasses ReFlowSET across all five reported quality metrics, including FID (19.1 $\rightarrow$ 15.5) and DISTS (0.231 $\rightarrow$ 0.212), requiring only two NFEs instead of 100. It also achieves the best FID and DISTS among the compared methods. The one-step LFA delivers competitive results (FID 17.4 and DISTS 0.213) with a single NFE. Relative to ReFlowSET, the two- and one-step models enable approximately $38\times$ and $76\times$ faster inference, respectively. Fig.~\ref{fig:qualitative} shows more coherent urban layouts, better-defined building geometry, and finer roof structures, with clearer road and building boundaries.

\noindent\textbf{SAR2Opt.}
Both low-step models outperform ReFlowSET on all four non-FID metrics. In particular, DISTS decreases from 0.185 to 0.182 and 0.181 with one and two steps, respectively. However, their FID scores remain higher than that of ReFlowSET (66.3), reaching 80.3 and 76.9, respectively. We hypothesize that the higher FID partly stems from the data sensitivity of pixel-space adaptation, consistent with the data-scaling observations of L2P~\cite{chen2026l2p}. In particular, SAR2Opt provides substantially fewer training pairs (1,450 vs. 16,001) at a higher resolution than QXS-SAROPT. This limited supervision may also explain the degradation from EPT to SPA, as the available data may be insufficient to fully adapt the SAR conditioning pathway after removing its pretrained VAE encoder. Qualitatively, Fig.~\ref{fig:qualitative} shows clearer building contours, finer textures, and natural vegetation patterns.

\subsection{Ablation Studies}
\label{sec:ablations}
\begin{table}[!t]
\centering
\footnotesize

\caption{\textbf{EPT training ablations on QXS-SAROPT.}
(a) Training design; (b) cumulative curriculum stages.}
\label{tab:phase_a_ablation}
\setlength{\tabcolsep}{1.5pt}
\renewcommand{\arraystretch}{1.14}
\begin{tabularx}{\columnwidth}{@{}c>{\centering\arraybackslash}Xc*{5}{c}@{}}
\toprule
\multicolumn{3}{@{}l}{\textbf{Configuration}}
& FID$\downarrow$ & DISTS$\downarrow$ & LPIPS$\downarrow$
& SSIM$\uparrow$ & PSNR$\uparrow$\\
\midrule
\rowcolor{gray!10}
\multicolumn{8}{@{}l}{\textit{(a) Training design}}\\
Init. & Sampler & Staged & & & & & \\
\xmark & LN$\to$U & \cmark
&20.6&.235&.539&.336&16.10\\
\cmark & U & \cmark
&19.4&\underline{.229}&\underline{.520}&\underline{.358}&\textbf{17.10}\\
\cmark & LN & \cmark
&\underline{18.3}&.238&.527&.351&15.92\\
\cmark & LN$\to$U & \xmark
&20.0&.232&.523&.354&17.01\\
\lfarow
\cmark & \textbf{LN$\to$U} & \cmark
&\textbf{17.9}&\textbf{.225}&\textbf{.516}&\textbf{.360}&\underline{17.02}\\
\midrule
\rowcolor{gray!10}
\multicolumn{8}{@{}l}{\textit{(b) Curriculum stages}}\\
\multicolumn{3}{@{}l}{Stage I}
&49.0&.275&.532&\textbf{.420}&\textbf{17.45}\\
\multicolumn{3}{@{}l}{Stage I $\to$ II}
&22.7&.236&\textbf{.502}&.403&17.35\\
\lfarow
\multicolumn{3}{@{}l}{Stage I $\to$ II $\to$ III}
&\textbf{17.9}&\textbf{.225}&.516&.360&17.02\\
\bottomrule
\end{tabularx}
\vspace{-10pt}
\end{table}

\noindent\textbf{EPT training ablation.}
 Table~\ref{tab:phase_a_ablation} (a) compares random and pretrained ReFlowSET initialization, logit-normal (LN) and uniform (U) timestep sampling, and staged versus all-at-once application of the shared curriculum objectives (Eq.~\ref{eq:stage_objectives}). Pretrained ReFlowSET initialization improves all five metrics over random initialization (FID: 20.6 to 17.9). The LN$\to$U schedule achieves the best FID, DISTS, and LPIPS by emphasizing noise-dominated states early and covering diverse timesteps later. Staged supervision further reduces FID from 20.0 to 17.9 compared with applying all losses from the beginning. Table~\ref{tab:phase_a_ablation} (b) further examines the cumulative effects of the three-stage curriculum. Stage II introduces perceptual supervision, reducing FID from 49.0 to 22.7 and DISTS from 0.275 to 0.236. Stage III adds adversarial refinement, reducing FID to 17.9 and DISTS to 0.225.

\begin{table}[!t]
\centering
\footnotesize
\caption{\textbf{Ablations on conversion-strategy and SPA-supervision for QXS-SAROPT.}
(a) Conversion strategies; (b) SPA supervision components.
For Joint, $\checkmark$ denotes joint optimization of both modality pathways, while $\triangle$ denotes DiT backbone optimization with frozen VAE interfaces. For Unc., $\checkmark$ denotes pixel-wise PUFM and $\triangle$ denotes latent-space uncertainty supervision~\cite{do2026c}.}
\label{tab:phase_b_ablation}
\setlength{\tabcolsep}{0.8pt}
\renewcommand{\arraystretch}{1.14}
\begin{tabularx}{\columnwidth}{
@{}
>{\raggedright\arraybackslash}p{0.235\columnwidth}
>{\centering\arraybackslash}p{0.06\columnwidth}
>{\centering\arraybackslash}p{0.07\columnwidth}
>{\centering\arraybackslash}p{0.075\columnwidth}
*{5}{>{\centering\arraybackslash}X}
@{}}
\toprule
\textbf{Configuration}
& CPD & Joint & Unc.
& FID$\downarrow$ & DISTS$\downarrow$
& LPIPS$\downarrow$ & SSIM$\uparrow$ & PSNR$\uparrow$\\
\midrule
\rowcolor{gray!10}
\multicolumn{9}{@{}l}{\textit{(a) conversion strategy}}\\
Latent FT
& -- & $\triangle$ & $\triangle$
& 25.4 & .237 & .515 & .358 & 16.75\\
Direct Conv.
& -- & $\checkmark$ & --
& 19.5 & .235 & .531 & .352 & 16.58\\
Frozen SAR
& -- & -- & $\checkmark$
& \textbf{16.4} & \textbf{.219} & \textbf{.509} & \textbf{.362} & 17.00\\
\midrule
\rowcolor{gray!10}
\multicolumn{9}{@{}l}{\textit{(b) SPA supervision}}\\
Baseline
& -- & -- & --
& 17.5 & .226 & .518 & .360 & 16.91\\
+ CPD
& $\checkmark$ & -- & --
& 17.2 & .223 & .513 & \textbf{.362} & \textbf{17.09}\\
+ Joint
& $\checkmark$ & $\checkmark$ & --
& 17.2 & \underline{.221} & \underline{.511} & \underline{.361} & \underline{17.07}\\
\lfarow
+ PUFM (Ours)
& $\checkmark$ & $\checkmark$ & $\checkmark$
& \underline{16.8} & \underline{.221} & \textbf{.509} & \textbf{.362} & 17.00\\
\bottomrule
\end{tabularx}
\vspace{-10pt}
\end{table}

\begin{figure}[t]
    \centering
    \includegraphics[width=\columnwidth]{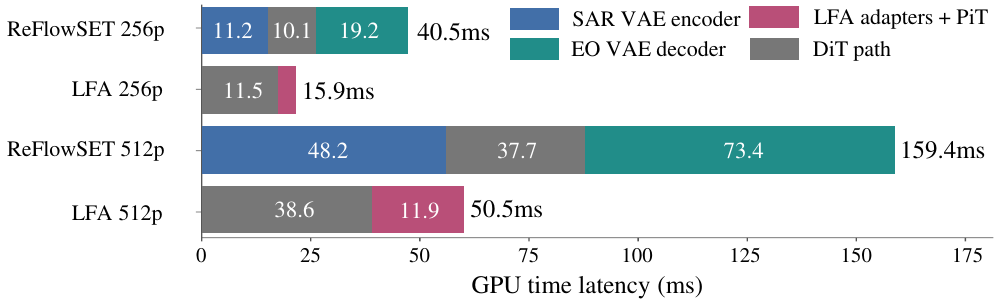}
    
    \caption{\textbf{Single-pass latency breakdown at
        $256\times256$ and $512\times512$.}
        VAE operations account for about 75\% of ReFlowSET's
        latency, with SAR encoding alone (48.2\,ms) nearly
        matching the entire VAE-free LFA forward pass
        (50.9\,ms) at $512\times512$.}

    \vspace{-12pt}
\label{fig:1nfe_compare}
\end{figure}

\noindent\textbf{Conversion strategy ablation.}
Table~\ref{tab:phase_b_ablation} (a) compares three alternative conversion strategies under a fixed 120k-update budget. 
(i) \textbf{\textit{Latent FT}} retains the frozen ReFlowSET VAE interfaces and sequentially trains using the EPT and SPA curriculum objectives for 60k updates each. Its FID of 25.4, compared with 16.8 for full SPA, indicates that applying the same curriculum to the latent generator alone does not match the performance of the full pixel-space adaptation framework.
(ii) \textbf{\textit{Direct Conv}} simultaneously replaces both SAR and EO VAE encoders with LPAs and the EO VAE decoder with PiT, training with $\mathcal L_{\mathrm{curr}}^{\mathrm{EPT}}(k)$ for 40k updates per stage. Since Direct Conv bypasses EPT, EPT-based CPD is unavailable.
Nevertheless, its FID of 19.5 is higher than both the
CPD-free SPA baseline (17.5) and SPA with CPD and joint
optimization (17.2), supporting staged conversion.
(iii) \textbf{\textit{Frozen SAR}} starts from EPT and optimizes $\mathcal L_{\mathrm{curr}}^{\mathrm{SPA}}(k)$ for 60k additional updates while keeping the SAR conditioning pathway frozen.
It achieves slightly better translation quality in FID and DISTS, potentially benefiting from the rich pretrained SAR representations preserved by the frozen VAE-based conditioning pathway. However, retaining the SAR VAE encoder introduces fixed inference overhead, which becomes a major bottleneck in the low-step regime (see Fig.~\ref{fig:1nfe_compare}).
In contrast, full SPA eliminates this remaining VAE dependency while maintaining competitive SET quality, providing an efficient foundation for subsequent low-step adaptation.

\noindent\textbf{SPA supervision ablation.}
As shown in Table~\ref{tab:phase_b_ablation} (b), adding CPD to the baseline reduces FID from 17.5 to 17.2 while also improving DISTS and LPIPS. Joint optimization of both modality pathways further reduces DISTS to 0.221. Finally, adding PUFM achieves the lowest FID of 16.8 while maintaining competitive perceptual quality.

\begin{table}[!t]
\centering
\footnotesize
\caption{\textbf{Low-step training ablations on QXS-SAROPT.}
Results are reported for independently trained one- and two-step students.}
\label{tab:phase_c_ablation}
\setlength{\tabcolsep}{2.0pt}
\renewcommand{\arraystretch}{1.15}

\resizebox{\columnwidth}{!}{%
\begin{tabular}{@{}lcrrrrr@{}}
\toprule
\textbf{Method} & NFE
& FID$\downarrow$
& DISTS$\downarrow$
& LPIPS$\downarrow$
& SSIM$\uparrow$
& PSNR$\uparrow$\\

\midrule
\multirow{2}{*}{ReFlowSET + DMD2~\cite{yin2024improved}}
& 1 & 27.2 & .247 & .566 & .308 & 15.20\\
& 2 & 21.5 & .244 & .560 & .322 & 15.20\\

\multirow{2}{*}{Generative FM~\cite{xu2026pointdit}}
& 1 & 32.0 & .238 & .517 & .295 & 16.28\\
& 2 & 26.5 & .235 & .518 & .302 & 16.42\\

\multirow{2}{*}{LFA w/o teacher}
& 1 & 19.2 & .215 & .499 & .357 & 16.94\\
& 2 & \underline{17.3} & \textbf{.212} & \textbf{.495} & \textbf{.369} & \underline{17.14}\\

\midrule
\lfarow
& 1 & 17.4 & \underline{.213} & .498 & \underline{.365} & 17.12\\
\lfarow
\multirow{-2}{*}{\textbf{Full LFA (ours)}}
& 2 & \textbf{15.5} & \textbf{.212} & \underline{.496} & \textbf{.369} & \textbf{17.17}\\

\bottomrule
\end{tabular}%
}
\end{table}

\noindent\textbf{Low-step training ablation.}
Table~\ref{tab:phase_c_ablation} evaluates low-step training strategies and the effectiveness of teacher supervision. Direct DMD2 distillation~\cite{yin2024improved} from latent ReFlowSET without EPT/SPA/LFA yields FIDs of 27.2/21.5 for one-/two-step generation, whereas our LFA achieves lower FIDs of 17.4/15.5. Without teacher supervision, SPA-initialized LFA achieves FIDs of 19.2/17.3, which further improve to 17.4/15.5 with auxiliary endpoint supervision from the frozen SPA EMA teacher. These results suggest that SPA serves not only as an effective initialization for low-step adaptation but also as a valuable source of auxiliary supervision. Following PointDiT~\cite{xu2026pointdit}, we also evaluate a Generative FM variant that applies the LFA objective with probability 0.1 and the SPA Stage-3 objective with probability 0.9. Its higher FIDs (32.0/26.5) suggest that fixed-timestep regression using only the LFA objective is more effective than stochastically alternating between the LFA and SPA objectives.

\ifdefined\FloatBarrier\FloatBarrier\fi

\section{Conclusion}
\label{sec:conclusion}

We presented SPACE-SET, a staged framework that converts pretrained latent SAR-to-EO generators into VAE-free pixel-space models through EPT, SPA, and LFA. Progressive conversion preserves pretrained cross-modal representations while enabling pixel-level supervision. Our two-step model achieves state-of-the-art results on most quality metrics, while our one-step model maintains competitive quality with 76.07$\times$ faster inference than ReFlowSET. These results highlight the potential of efficient SET for large-scale remote-sensing analysis and downstream applications.

\clearpage
\section*{LLM Usage Statement}
ChatGPT (OpenAI) was used to obtain reviewer-style feedback on the manuscript and to assist with language editing, mathematical-notation consistency, and the organization of the main paper and supplementary material. It also assisted with refining author-provided code and improving the presentation and consistency of experimental protocols and analyses based on author-supplied settings and results. All research ideas, scientific claims, experimental results, and visual materials remain the work and responsibility of the authors.

{
    \small
    \bibliographystyle{ieeenat_fullname}
    \bibliography{main}
}

\providecommand{\mainref}[1]{\ref{#1}}
\providecommand{\NA}{\textemdash}
\providecommand{\GeoSuppOurRow}{\rowcolor{oursred}}
\providecommand{\GeoSuppTableStyle}{%
  \footnotesize\setlength{\tabcolsep}{3pt}%
  \renewcommand{\arraystretch}{1.13}}
\ifdefined\ifSuppDraft\else
  \newif\ifSuppDraft
  \SuppDrafttrue
\fi
\providecommand{\SuppDraftNote}[1]{%
  \ifSuppDraft\par\smallskip\noindent
  \begingroup\setlength{\fboxsep}{4pt}%
  \colorbox{yellow!12}{\parbox{\dimexpr\linewidth-8pt\relax}{%
  \small\textbf{Draft status.} #1}}%
  \endgroup\par\smallskip\fi}
\providecommand{\SuppAsset}[3]{%
  \IfFileExists{#1}{\includegraphics[width=\linewidth]{#1}}{%
  \begingroup\setlength{\fboxsep}{7pt}%
  \fbox{\begin{minipage}[c][#2][c]{\dimexpr\linewidth-2\fboxsep-2\fboxrule\relax}
  \centering\small\textcolor{black!65}{#3}
  \end{minipage}}\endgroup}}

\clearpage
\setcounter{page}{1}
\renewcommand{\thepage}{S\arabic{page}}
\appendix
\setcounter{equation}{0}
\renewcommand{\theequation}{S\arabic{equation}}
\setcounter{table}{0}
\renewcommand{\thetable}{S\arabic{table}}
\setcounter{figure}{0}
\renewcommand{\thefigure}{S\arabic{figure}}
\setcounter{algorithm}{0}
\renewcommand{\thealgorithm}{S\arabic{algorithm}}
\ifdefined\theHequation
  \renewcommand{\theHequation}{supp.eq.\arabic{equation}}
\fi
\ifdefined\theHtable
  \renewcommand{\theHtable}{supp.tab.\arabic{table}}
\fi
\ifdefined\theHfigure
  \renewcommand{\theHfigure}{supp.fig.\arabic{figure}}
\fi
\ifdefined\theHalgorithm
  \renewcommand{\theHalgorithm}{supp.alg.\arabic{algorithm}}
\fi
\raggedbottom
\setcounter{dbltopnumber}{3}
\renewcommand{\dbltopfraction}{0.95}
\renewcommand{\textfraction}{0.04}
\makeatletter
\setlength{\@dblfptop}{0pt}
\setlength{\@dblfpsep}{12pt}
\ifdefined\theHALG@line
  \renewcommand*{\theHALG@line}{supp.alg.\arabic{algorithm}.line.\arabic{ALG@line}}
\fi
\makeatother

\twocolumn[{
\begin{center}
{\Large\bfseries
\method{}: Staged Pixel-Space Adaptation of Conditioning\\
and EO Generation for Low-Step SAR-to-EO Translation\par}
\vspace{10pt}
{\large\bfseries Supplementary Material\par}
\end{center}

\begin{minipage}{\textwidth}
\centering
\includegraphics[width=\linewidth]{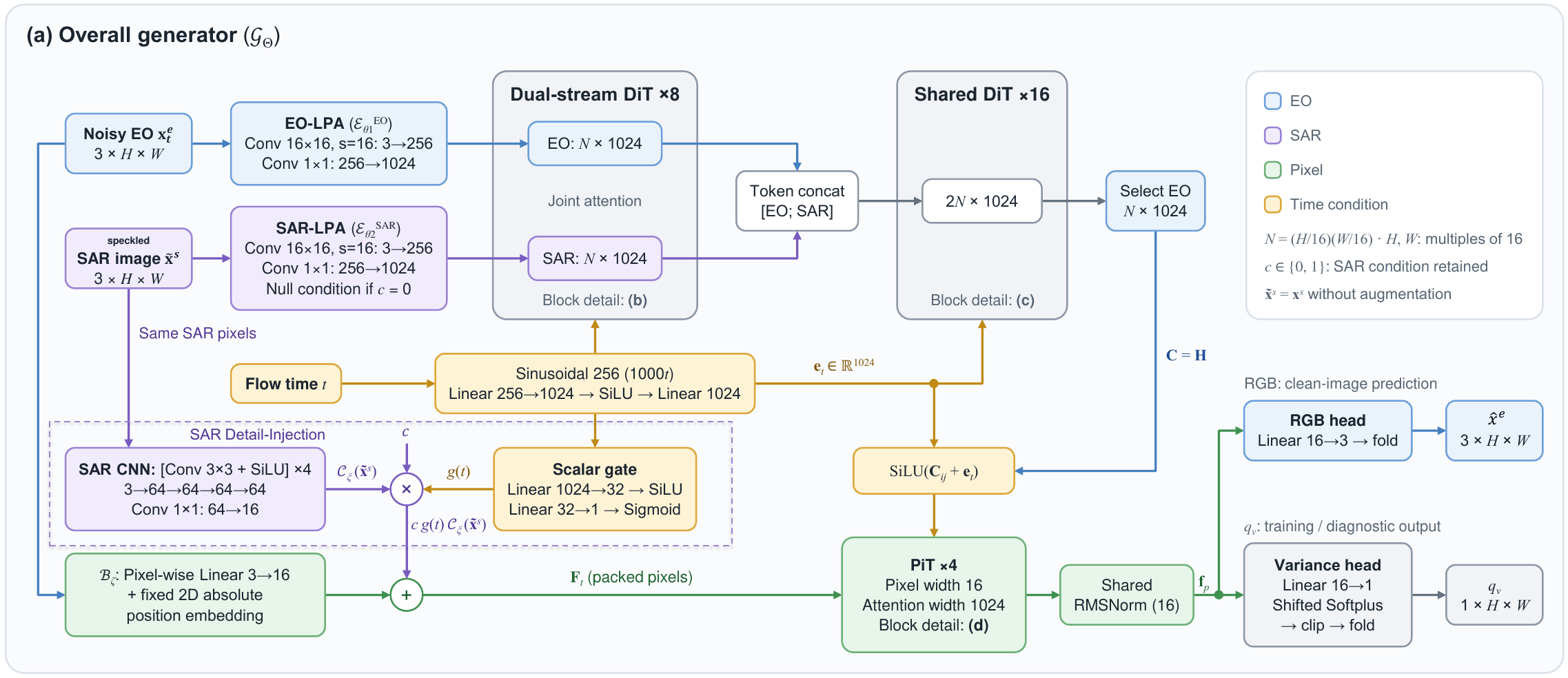}

\captionof{figure}{\textbf{Overall generator architecture of \method{}.}
The fully converted SPA generator uses SAR and EO pixel adapters, dual-stream and shared DiT blocks, and a PiT pixel decoder. The dashed box encloses the timestep-gated SAR detail-injection
branch used during training and evaluation.
Figure~\ref{fig:supp_sar_gate_analysis} analyzes its learned gate. We use the variance head during training and to visualize
predicted uncertainty in Fig.~\ref{fig:supp_pufm_uncertainty}.
We omit it during standard inference.}
\vspace{10pt}
\label{fig:supp_overall_architecture}
\end{minipage}
}]

\noindent
Table~\ref{tab:supp_overview} provides an overview of the supplementary material.
Sections~\ref{supp:architecture}--\ref{supp:ufm} describe the
generator architecture and ablation protocols, then examine
timestep-dependent supervision, SAR conditioning features,
and pixel-wise uncertainty.
Sections~\ref{supp:training}--\ref{supp:limitations} cover
training and low-step inference, followed by qualitative
comparisons, failure cases, and limitations.
Unless otherwise specified, ablations use QXS-SAROPT and
the phase-specific inference settings in
Sec.~\mainref{sec:exp_setup}.

\begin{table}[!ht]
\centering
\footnotesize
\caption{\textbf{Overview of the supplementary material.}}
\label{tab:supp_overview}
\setlength{\tabcolsep}{4pt}
\renewcommand{\arraystretch}{1.12}
\begin{tabularx}{\columnwidth}{l>{\raggedright\arraybackslash}X}
\toprule
\rowcolor{oursred}
\textbf{Section} & \textbf{Contents}\\
\midrule
\ref{supp:architecture} & Generator, decoder comparison, and SAR detail-injection gate analysis\\
\ref{supp:ablation_protocols} & Protocols for Tables~\mainref{tab:phase_a_ablation}--\mainref{tab:phase_c_ablation} and a comparison of conversion orders\\
\ref{supp:conversion_diagnostics} & Timestep weighting, its exponent, and EO-conditioned SAR features\\
\ref{supp:ufm} & PUFM gradients, sensitivity to $\beta$, and uncertainty maps\\
\ref{supp:training} & SAR augmentation ablation, losses, algorithms, and learning curves\\
\ref{supp:lfa} & Inference-step sweep, low-step students, and larger-patch adaptation\\
\ref{supp:qualitative} & Evaluation metrics, qualitative comparisons, and failure cases\\
\ref{supp:limitations} & SAR-specific priors, training complexity, and student design\\
\bottomrule
\end{tabularx}
\end{table}

\section{Generator Architecture}
\label{supp:architecture}

\subsection{Detailed Generator Architecture}
\label{supp:pixel_interfaces}

Figures~\ref{fig:supp_overall_architecture} and~\ref{fig:supp_block_details} illustrate the fully converted generator and its constituent blocks. Each Lightweight Pixel Adapter (LPA) uses a $16\times16$ convolution with stride 16 ($3\to256$ channels), followed by a $1\times1$ convolution ($256\to1024$), to produce one token per non-overlapping image patch. For an $H\times W$ input, each stream contains $N=N_s=(H/16)(W/16)$ tokens with feature dimension $D_h=1024$, matching the CPD notation in Eq.~\mainref{eq:kd_loss}. Eight dual-stream DiT blocks process the EO and SAR tokens with modality-specific parameters and joint attention; 16 shared DiT blocks then process their concatenation. The EO tokens condition a four-block PiT decoder, which maintains 16-dimensional features at each pixel within the patches and produces a clean EO prediction through an RGB head. A separate training-only variance head operates on the final pixel features (Sec.~\ref{supp:ufm}).

\begin{figure*}[t]
\centering
\includegraphics[width=\textwidth]{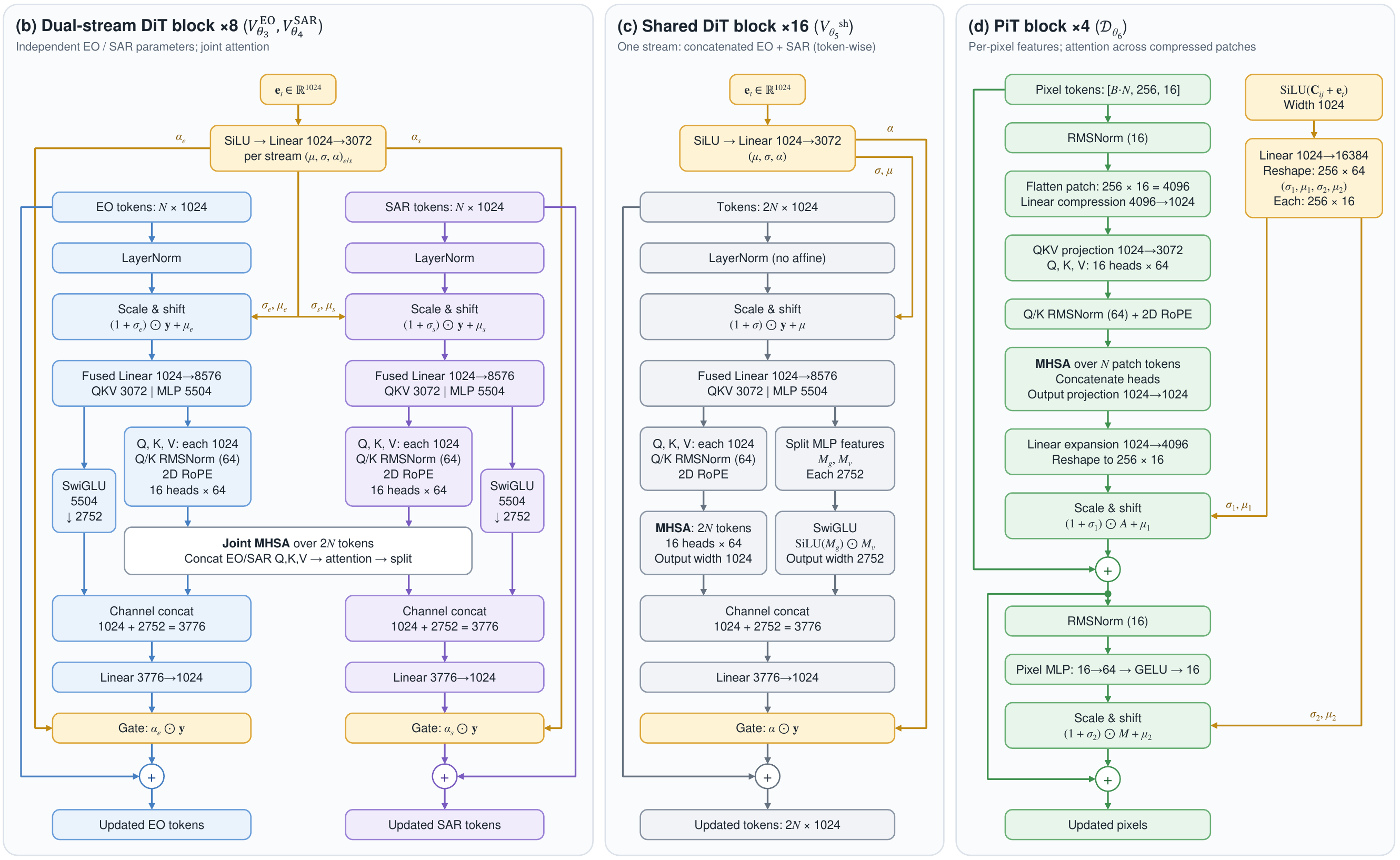}
\caption{\textbf{Detailed generator blocks.} (b) Dual-stream DiT block with modality-specific transformations and joint attention. (c) Shared DiT block processing concatenated EO and SAR tokens. (d) PiT block using attention across patches and pixel-wise refinement within each patch. Panel (a), the overall generator, appears in Fig.~\ref{fig:supp_overall_architecture}.}
\label{fig:supp_block_details}
\vspace{-5px}
\end{figure*}

\subsection{Pixel Decoder Comparison}
\label{supp:decoder_ablation}

We compare PiT~\cite{yu2026pixeldit} with a JiT-style linear
patch head, DeCo~\cite{Ma_2026_CVPR}, and DiP~\cite{chen2026dip}
under the same EPT training and evaluation protocol,
changing only the decoder.
The linear head directly maps each EO token to a
$16\times16$ RGB patch, whereas PiT combines attention
across patches with pixel-wise local refinement
(Fig.~\ref{fig:supp_block_details}).

Table~\ref{tab:supp_decoder_ablation} shows that the JiT
linear head reduces decoding latency to just $0.01$\,ms,
but its perceptual quality remains limited.
Both DeCo and DiP improve FID and DISTS over JiT
at the expense of increased decoding latency.
PiT achieves the lowest FID, DISTS, and LPIPS and the
highest SSIM among the evaluated pixel decoders,
while maintaining a PSNR close to JiT
($17.02$ vs.\ $17.06$\,dB).
Although PiT has more parameters and increases decoding
latency to $3.41$\,ms, its decoding cost remains
substantially below the VAE reference at $19.49$\,ms.
We therefore adopt PiT for its improved perceptual
quality at a moderate decoding cost.

\begin{table}[!t]
\centering\GeoSuppTableStyle
\setlength{\tabcolsep}{1.5pt}

\caption{\textbf{Pixel decoder comparison on QXS-SAROPT.}
Pixel-decoder variants share the EPT training and evaluation
protocol. Parameters and latency are measured for the
decoder only under identical conditions.
The gray VAE row provides a reference comparison.}
\label{tab:supp_decoder_ablation}

\renewcommand{\tabularxcolumn}[1]{m{#1}}

\begin{tabularx}{\columnwidth}{
    @{}l*{7}{>{\centering\arraybackslash}X}@{}
}
\toprule
Dec.
& \shortstack{Params\\(M)}
& \shortstack{Latency$\downarrow$\\(ms)}
& FID$\downarrow$
& DISTS$\downarrow$
& LPIPS$\downarrow$
& SSIM$\uparrow$
& PSNR$\uparrow$\\
\midrule

\textcolor{gray}{VAE~\cite{kingma2014vae}}
& \textcolor{gray}{49.62}
& \textcolor{gray}{19.49}
& \textcolor{gray}{\NA}
& \textcolor{gray}{\NA}
& \textcolor{gray}{\NA}
& \textcolor{gray}{\NA}
& \textcolor{gray}{\NA}\\
\midrule

JiT~\cite{li2026back}  & 0.79 & 0.01 & 20.6 & .234 & .524 & .359 & \textbf{17.06}\\
DeCo~\cite{Ma_2026_CVPR} & 8.42 & 1.76 & 19.8 & .233 & .524 & .358 & 17.00\\
DiP~\cite{chen2026dip}  & 6.90 & 1.45 & 20.4 & .229 & .522 & .357 & 17.00\\

\GeoSuppOurRow
PiT~\cite{yu2026pixeldit} & 117.54 & 3.41
& \textbf{17.9}
& \textbf{.225}
& \textbf{.516}
& \textbf{.360}
& 17.02\\

\bottomrule
\end{tabularx}
\end{table}

\subsection{Timestep-Gated SAR Detail Injection}
\label{supp:sare}

The timestep-gated SAR detail-injection branch introduced
in Sec.~\mainref{sec:spa} provides a direct path from SAR
pixels to the PiT decoder. We retain this branch during
training and evaluation of the reported model.
Overall translation quality was comparable with and without
the branch.
We analyze its learned gate $g(t)$ to examine how the direct
SAR residual is weighted across flow times
(Fig.~\ref{fig:supp_sar_gate_analysis}).

A lightweight convolutional network $\mathcal C_\xi$ maps
the SAR input to a residual that is added to the noisy-EO
pixel embeddings before PiT:
\begin{equation}
\mathbf F_t
=\mathcal B_\zeta(\mathbf x_t^e)
+c\,g(t)\,\mathcal C_\xi(\tilde{\mathbf x}^s).
\label{eq:supp_sar_residual}
\end{equation}
Here $\mathcal B_\zeta$ is the pixel-wise $3\to16$
embedding with fixed 2D positional embeddings, and
$c\in\{0,1\}$ indicates whether SAR conditioning is retained.
This embedding is distinct from the REPA projector
$\mathcal P_\omega$ in Eq.~\mainref{eq:repa}.
The SAR network uses four $3\times3$ convolution--SiLU
blocks with channel widths $3\to64\to64\to64\to64$,
followed by a $1\times1$ projection to 16 channels.
A two-layer MLP (1024$\to$32$\to$1), with SiLU and
sigmoid activations, predicts $g(t)\in(0,1)$ from the
timestep embedding.
Dropping SAR conditioning replaces the SAR
tokens with the null condition and sets $c=0$.

\begin{figure}[!t]
\centering
\includegraphics[width=\columnwidth]
{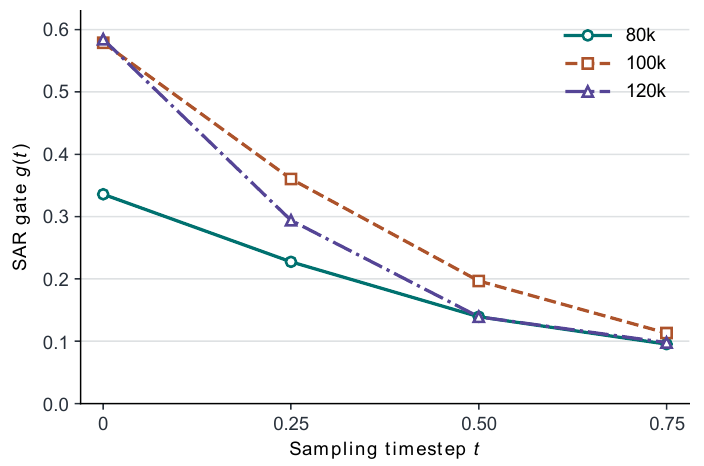}
\caption{\textbf{Learned gate of the SAR detail-injection branch during SPA.}
Raw checkpoints at 80k, 100k, and 120k cumulative updates
correspond to the ends of SPA Stages~I, II, and III.
The Stage-I gate already assigns larger values to earlier
flow times, before output-level perceptual and adversarial
losses are activated during SPA.
Later-stage checkpoints exhibit greater early--late contrast.
The gate scales the direct local-feature injection pathway;
SAR-token conditioning remains available through the backbone.}
\label{fig:supp_sar_gate_analysis}
\end{figure}

\paragraph{Learned gates during SPA.}
\label{supp:sar_gate_analysis}
At the 80k checkpoint, which marks the end of SPA Stage~I,
the learned gate is largest at $t=0$ and decreases across
the evaluated flow times (Fig.~\ref{fig:supp_sar_gate_analysis}).
Thus, even before perceptual and adversarial losses are
introduced during SPA, the decoder gives greater weight to
local SAR features when the EO input is dominated by noise.
The contrast between early and late gate values becomes more
pronounced at the 100k and 120k checkpoints.

This pattern suggests that the direct SAR pathway helps
establish scene structure at early flow times. At later
times, the gate attenuates the local SAR residual as the
EO state becomes cleaner, while SAR information remains
available through the backbone at every conditioned step.

\section{Additional Details on the Ablation Studies}
\label{supp:ablation_protocols}

We use the QXS-SAROPT split and inference settings in
Sec.~\mainref{sec:exp_setup}. Training budgets count generator updates.

\subsection{EPT Initialization, Sampling, and Curriculum}
\label{supp:ept_protocol}

\noindent\textbf{Training design.}
To test whether EPT benefits from latent-to-pixel transfer as in
L2P~\cite{chen2026l2p}, we compare pretrained ReFlowSET weights with
random initialization in Table~\mainref{tab:phase_a_ablation}(a). To test the timestep schedule,
we compare the logit-normal-to-uniform switch with each sampler used
alone. To test cumulative supervision, we keep this switch and compare
adding perceptual and adversarial terms in stages with enabling the
full objective from the first update. Each comparison varies one
choice and evaluates the model at 60k updates.

\noindent\textbf{Contributions of the curriculum stages.}
To assess how each added objective affects final quality, panel~(b)
compares Stage~I alone, Stage~I plus perceptual supervision, and the
full curriculum with adversarial supervision added last. Each run
divides 60k updates equally among its stages (60k, 30k, or 20k per
stage, respectively). The 20k and 40k points in
Fig.~\ref{fig:supp_validation_curves} are checkpoints of the full run.

\subsection{Conversion Strategy and SPA Supervision}
\label{supp:spa_protocol}

\noindent\textbf{Conversion strategy.}
To test whether the curriculum alone can match pixel-space adaptation,
\emph{Latent FT} applies it to latent ReFlowSET. To test whether simultaneous
conversion can match EPT followed by SPA, \emph{Direct Conv.} replaces both
input interfaces and the EO decoder at initialization and trains both
pathways jointly. To test whether SPA can remove the low-step SAR VAE bottleneck
while preserving translation quality, \emph{Frozen SAR} retains the SAR
VAE and continues EPT with PUFM and no CPD. At 120k updates
(Table~\mainref{tab:phase_b_ablation}), full SPA matches Frozen SAR on
LPIPS, SSIM, and PSNR at the reported precision, with slightly higher
FID and DISTS.

\noindent\textbf{SPA supervision.}
To measure what CPD, joint optimization, and PUFM each add, panel~(b)
introduces them successively over 60k updates from the same 60k EPT
checkpoint. \emph{Baseline} freezes the EO pathway and uses standard
FM. \emph{+CPD} adds SAR-token supervision from the frozen EPT EMA
teacher. \emph{+Joint} also unfreezes the EO pathway. \emph{+PUFM} retains both changes and replaces FM with
PUFM ($\beta=1$). SAR augmentation is tested separately in
Table~\ref{tab:supp_spa_augmentation}.

\subsection{Order of EO and SAR Conversion}
\label{supp:conversion_order}


We compare two orders for converting the SAR and EO pathways
to pixel space under the same total budget of 120k updates.
In the EO-first schedule, EPT retains the frozen SAR pathway.
During the subsequent joint adaptation, the student matches
the SAR tokens of the frozen EPT EMA teacher.
In the SAR-first schedule, the EO pathway is frozen during
the first phase. During joint adaptation, the student matches
the EO tokens of its frozen first-phase teacher.

Table~\ref{tab:supp_conversion_order} reports the first-phase
checkpoints at 60k updates and the jointly adapted models at
120k updates. The comparison therefore includes the teacher
and the distillation target produced by each conversion order.

\begin{table}[!ht]
\centering\GeoSuppTableStyle
\caption{\textbf{Pathway conversion order on QXS-SAROPT.}
The EO-first checkpoints are EPT and full SPA from
Tables~\mainref{tab:phase_a_ablation}(b)
and~\mainref{tab:phase_b_ablation}(b), respectively.}
\label{tab:supp_conversion_order}
\setlength{\tabcolsep}{2.3pt}
\begin{tabular*}{\columnwidth}{@{\extracolsep{\fill}}lrrrrr@{}}
\toprule
Order & FID$\downarrow$ & DISTS$\downarrow$ & LPIPS$\downarrow$
& SSIM$\uparrow$ & PSNR$\uparrow$\\
\midrule
\rowcolor{gray!10}
\multicolumn{6}{@{}l}{\emph{First-phase teacher (60k updates)}}\\
SAR-first      & 20.5 & .244 & .538 & .351 & 16.65 \\
EO-first (EPT) & \textbf{17.9} & \textbf{.225} & \textbf{.516} & \textbf{.360} & \textbf{17.02} \\
\midrule
\rowcolor{gray!10}
\multicolumn{6}{@{}l}{\emph{Joint model (120k total updates)}}\\
SAR-first      & 17.3 & .232 & .527 & .351 & 16.63 \\
EO-first (SPA) & \textbf{16.8} & \textbf{.221} & \textbf{.509} & \textbf{.362} & \textbf{17.00} \\
\bottomrule
\end{tabular*}
\end{table}

At 60k updates, the EO-first EPT teacher outperforms the SAR-first
teacher on all five metrics, with lower FID (17.9 vs.\ 20.5) and
DISTS (0.225 vs.\ 0.244). After a further 60k updates of joint
adaptation, the FID gap narrows (16.8 vs.\ 17.3), but the EO-first
SPA model retains its lead across all five metrics. In particular,
SPA achieves lower DISTS (0.221 vs.\ 0.232) and LPIPS
(0.509 vs.\ 0.527), along with higher SSIM and PSNR.

\subsection{Low-Step Training Strategies}
\label{supp:lfa_protocol}

To test whether direct distillation of latent ReFlowSET can match LFA
at one or two steps, we train \emph{ReFlowSET + DMD2} using the official
implementation~\cite{yin2024improved}. Each student receives the same
10k generator updates as full LFA. Our two time-scale update rule
(TTUR) uses four discriminator updates per generator update, or 40k
per student.

To test whether SPA provides guidance beyond its role as the student's
starting point, we compare full LFA and \emph{LFA w/o teacher} at each
step count in Table~\mainref{tab:phase_c_ablation}. Both use the same
raw SPA initialization and paired-EO supervision. The teacher-free
variant removes the MAE and perceptual losses on the frozen SPA EMA
teacher's guided endpoint (Eq.~\ref{eq:supp_lfa_effective_weights}).

To test whether fixed-timestep regression is preferable for SET,
we compare LFA with \emph{Generative FM}, which mixes the LFA and SPA
Stage-III objectives. PointDiT~\cite{xu2026pointdit} found that
generative FM converged more slowly but reached lower validation error
than deterministic regression in monocular geometry estimation. In our extended runs, no Generative FM
checkpoint matched the corresponding 10k LFA student. Fixed-timestep
LFA showed signs of overfitting beyond 20k updates. Whether a larger
budget would close the gap remains open.

\section{Diagnostics of Pixel-Space Conversion}
\label{supp:conversion_diagnostics}

We first examine when perceptual and adversarial losses act on an EO
prediction during flow matching. We then evaluate what information the
SAR tokens retain after conversion by comparing them with the EPT
teacher and by predicting EO attributes from frozen features.

\subsection{Analysis of Timestep-Dependent Behavior}
\label{supp:timestep_analysis}
\label{supp:timestep_gradients}

\paragraph{Output-space gradient diagnostics.}
We examine $\ell_{\mathrm{FM}}$, $\ell_{\mathrm P}$, and
$\ell_{\mathrm{GAN}}$ through their gradients with respect
to the predicted clean EO image $\hat{\mathbf x}^e$.
For these diagnostics, $\ell_k$ includes its fixed base
coefficient, with $k\in\{\mathrm{FM},\mathrm P,\mathrm{GAN}\}$:
\begin{equation}
\mathbf v_k=\nabla_{\hat{\mathbf x}^e}\ell_k,\qquad
\rho_k=\Pr[\cos(\mathbf v_{\mathrm{FM}},\mathbf v_k)<0].
\label{eq:supp_output_gradient_diagnostics}
\end{equation}
At each fixed timestep, $\rho_k$ is estimated as the
fraction of evaluated samples with negative gradient cosine
relative to FM.
We disable the timestep thresholding and weighting applied
to perceptual and generator-adversarial supervision, while
retaining the FM residual normalization.
The GAN head is evaluated before adversarial training,
so its gradients describe the initial adversarial signal.

Figure~\ref{fig:supp_timestep_diagnostics} (bottom left)
shows that perceptual and generator-adversarial gradient
norms substantially exceed the FM norm at small $t$.
The bottom-right panel shows frequent GAN--FM directional
disagreement near $t=0$, while the combined perceptual
objective exhibits negative cosine less frequently.
These diagnostics indicate that appearance supervision
can introduce strong, potentially competing output-space
gradients at early flow times.
An early-time threshold directly excludes these appearance
gradients from the affected interval, limiting their
potential conflict with flow-based reconstruction.

\paragraph{Clean-image predictions.}
The top panel shows clean EO predictions from the interpolated
inputs $\mathbf x_t^e$ in Eq.~\mainref{eq:prelim_flow_path}.
At $t=0$--$0.3$, the predictions capture coarse scene
structure, while textures and boundaries remain blurred.
At $t=0.8$ and $0.9$, the inputs contain a larger contribution
from the paired EO image, and the predictions recover finer
detail.
This progression is consistent with the larger weight assigned
to the direct SAR residual at early timesteps
(Fig.~\ref{fig:supp_sar_gate_analysis}).
Together, these observations suggest complementary roles across
flow time, with SAR-guided recovery of scene structure in early,
noise-dominated states and refinement of EO appearance as the
state becomes cleaner.

Together, the gradient diagnostics and clean-image
predictions motivate the timestep thresholding and
weighting introduced in the main paper.
Thresholding suppresses potentially competing appearance
gradients at early timesteps, while weighting gives
appearance losses larger coefficients at later timesteps.
Through this allocation, we aim to promote stable
optimization and reinforce the complementary roles of
early spatial reconstruction and later appearance refinement.
The schedule and its empirical behavior are examined in
Sec.~\ref{supp:stage2_analysis}.

\begin{figure*}[!t]
\centering
\includegraphics[width=\textwidth]
{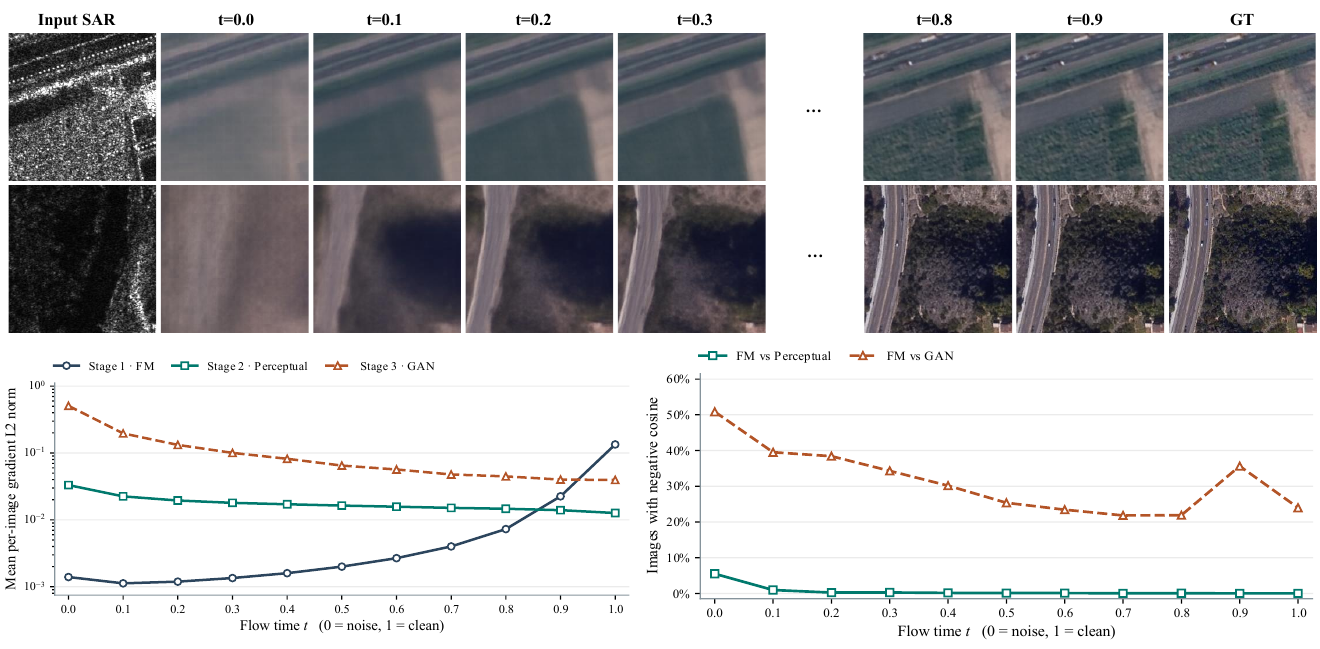}
\caption{\textbf{Timestep-dependent clean EO predictions and gradient analysis.}
Top: clean-image predictions become sharper at later flow times.
Bottom left: output-image gradient norms of FM, perceptual,
and generator-adversarial losses.
Bottom right: fraction of negative gradient cosine relative to FM.
Perceptual and adversarial timestep thresholding and weighting
are disabled, while FM residual normalization is retained.}
\label{fig:supp_timestep_diagnostics}
\vspace{-7px}
\end{figure*}

\subsection{Timestep Thresholding and Weighting}
\label{supp:stage2_analysis}
\label{supp:stage2_factorial}

The output-space diagnostics and clean-image predictions
in Sec.~\ref{supp:timestep_gradients} motivate a
timestep-dependent allocation of appearance supervision.
We detail the thresholding and weighting used in EPT/SPA
and evaluate their effects through EPT ablations.

\paragraph{Supervision schedule.}
To limit potentially competing early-time appearance
gradients, we exclude perceptual and generator-adversarial
losses for $t<\tau$.
For $t\geq\tau$, we apply $w(t)=d(t)^{-1}$ to increase
their coefficients toward the clean endpoint, where
$d(t)=\max(1-t,0.05)$.
The combined multiplier is
\begin{equation}
w_{\mathrm{eff}}(t)
=\ind[t\geq\tau]w(t)
=\frac{\ind[t\geq0.3]}{\max(1-t,0.05)}.
\label{eq:supp_effective_timestep_weight}
\end{equation}
Thresholding removes appearance-loss gradients at early
timesteps, while weighting gives these losses larger
coefficients at later timesteps.
FM/PUFM supervision remains active throughout EPT/SPA.
The perceptual objective is introduced in Stage~II,
and the generator-adversarial objective is added in
Stage~III; each uses this multiplier once activated.
By suppressing early-time appearance gradients and
progressively strengthening appearance supervision toward
the clean endpoint, the combined schedule encourages
a clearer separation between early-time spatial
reconstruction and late-time appearance refinement.
Figure~\ref{fig:supp_timestep_supervision} visualizes
the multiplier, excluding loss coefficients;
the main paper's conditional averaging is unchanged.

\begin{figure}[!t]
\centering
\includegraphics[width=\columnwidth]
{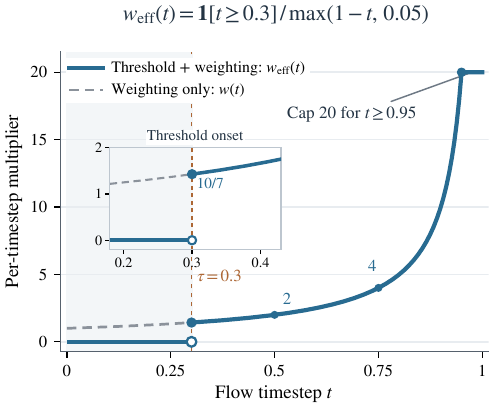}
\caption{\textbf{Combined timestep thresholding and weighting.}
The solid curve shows $w_{\mathrm{eff}}(t)$; the dashed curve
shows weighting without thresholding.
The combined multiplier is zero below $t=0.3$, increases
toward the clean endpoint, and is capped at $20$ for
$t\geq0.95$.
Loss coefficients are excluded.}
\label{fig:supp_timestep_supervision}
\end{figure}

\paragraph{Ablation results.}
The exponent $\gamma$ in $w(t)=d(t)^{-\gamma}$ controls how strongly
the appearance losses emphasize later timesteps.
In Table~\ref{tab:supp_stage2_factorial}, the default schedule
combines thresholding with $\gamma=1$ and achieves the lowest
FID, DISTS, and LPIPS.
With thresholding alone ($\gamma=0$), FID rises from 17.9 to 19.1,
DISTS from 0.225 to 0.226, and LPIPS from 0.516 to 0.521,
although SSIM and PSNR increase slightly (by 0.001 and
0.01\,dB, respectively).
With weighting alone ($\gamma=1$), appearance losses also act at
$t<0.3$. All five metrics worsen relative to the default.
FID rises to 21.7, SSIM falls from 0.360 to 0.319, and PSNR
falls from 17.02 to 16.12\,dB.

With thresholding retained, increasing $\gamma$ from 1 to 2
places more weight on late timesteps and worsens all five metrics.
For example, FID rises from 17.9 to 20.0 and PSNR falls from
17.02 to 16.65\,dB.

\begin{table}[!ht]
\centering\GeoSuppTableStyle
\caption{\textbf{Timestep thresholding and weighting in EPT.}
QXS-SAROPT results at 60k updates using four Euler steps
(six generator evaluations).
W denotes timestep weighting of the appearance losses, and T denotes
thresholding at $t\geq0.3$ (\cmark{} on, \xmark{} off).
Disabling W sets $\gamma=0$, so $w(t)=1$ wherever the losses are active.}
\label{tab:supp_stage2_factorial}
\begin{tabularx}{\columnwidth}{@{}ccc*{5}{>{\centering\arraybackslash}X}@{}}
\toprule
W & T & $\gamma$ & FID$\downarrow$ & DISTS$\downarrow$
& LPIPS$\downarrow$ & SSIM$\uparrow$ & PSNR$\uparrow$\\
\midrule
\cmark & \xmark & $1$ & 21.7 & .242 & .526 & .319 & 16.12\\
\xmark & \cmark & $0$ & 19.1 & .226 & .521 & \textbf{.361} & \textbf{17.03}\\
\cmark & \cmark & $2$ & 20.0 & .229 & .525 & .332 & 16.65\\
\GeoSuppOurRow
\cmark & \cmark & $1$ & \textbf{17.9} & \textbf{.225} & \textbf{.516} & .360 & 17.02\\
\bottomrule
\end{tabularx}
\end{table}
\FloatBarrier

\subsection{Distillation of EO-Conditioned SAR Representations}
\label{supp:sar_distillation}
\label{supp:sar_representations}

Converting the EO pathway first allows the EPT teacher to
process noisy EO images in pixel space while retaining the
pretrained VAE-based SAR pathway. During SPA, we use the frozen
60k-update EPT EMA checkpoint as the CPD teacher. Given the
same noisy EO image at the same flow time, the new pixel-space
SAR pathway is trained to reproduce the teacher's SAR features.

Figure~\ref{fig:supp_sar_distillation} compares how closely
full SPA and the Baseline in Table~\mainref{tab:phase_b_ablation}(b)
reproduce these EO-conditioned teacher features. We use SAR
tokens from the final dual-stream block ($L=8$) on 24 matched
QXS-SAROPT scenes.

\paragraph{Token values and spatial relationships.}
Panel~(a) of Fig.~\ref{fig:supp_sar_distillation},
comprising the two left plots, tracks raw student
checkpoints from their shared initialization.
Each checkpoint is compared with the frozen 60k EPT EMA
teacher at a fixed flow time of $t=0.25$.
The leftmost plot reports token MSE, which measures
differences in feature values at corresponding patch
locations.
The middle plot reports centered kernel alignment
(CKA)~\cite{kornblith2019similarity}, which measures
agreement in the relationships among patch features
within each scene.
We compute CKA over 256 spatial patches using
U-centered linear Gram matrices.

At 120k, SPA achieves a token MSE of $0.87$ versus $7.41$
for the Baseline, and a patch CKA of $0.838$ versus $0.487$.
Thus, SPA more closely reproduces both the teacher's
token values and its within-scene patch relationships.
The Baseline's MSE rises again after its initial decrease,
whereas SPA continues to improve its alignment with
the teacher.

\begin{figure*}[!t]
\centering
\includegraphics[width=\textwidth]
{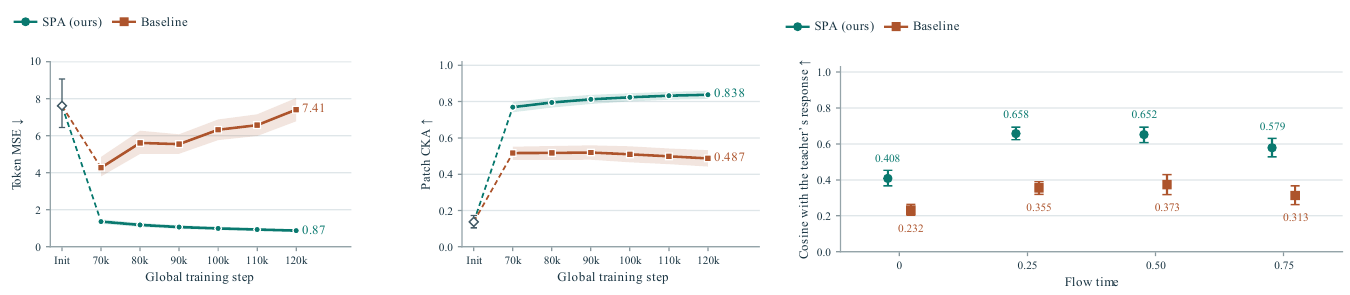}
\caption{\textbf{Distillation of EO-conditioned SAR representations.}
The frozen 60k EPT EMA model serves as the reference teacher.
\textbf{(a) Left two plots:}
token MSE (leftmost) and within-scene patch CKA (middle)
between raw student checkpoints and the teacher at $t=0.25$.
\textbf{(b) Rightmost plot:}
cosine agreement between teacher and 120k EMA student
SAR-token changes induced by two different EO noise draws.
The SAR observation, paired EO image, and timestep are
fixed within each comparison, and the two noise draws
are shared across models.
Higher cosine indicates a more teacher-like response
to the change in noisy EO input.
The Baseline configuration is specified in
Table~\mainref{tab:phase_b_ablation}(b).
All features are taken from SAR block 8;
shaded bands and error bars show 95\% scene-bootstrap intervals.}
\label{fig:supp_sar_distillation}
\end{figure*}

\paragraph{Responses to noisy EO inputs.}
Panel~(b) of Fig.~\ref{fig:supp_sar_distillation},
the rightmost plot, asks whether SPA also reproduces
how the teacher's SAR tokens respond to changes in
the noisy EO input.
Although the SAR observation is unchanged, its tokens
can change because they interact with EO tokens
in the dual-stream blocks.
We compare the 120k EMA SPA and Baseline checkpoints
with the same frozen 60k EPT EMA teacher.

For each scene and timestep, we fix the SAR observation
and paired EO image, and construct two noisy EO inputs:
\begin{equation}
\mathbf x_{t,a}^{e}
=(1-t)\boldsymbol\epsilon_a+t\mathbf x^e,
\qquad a\in\{1,2\},
\label{eq:supp_eo_response_inputs}
\end{equation}
where the same two Gaussian noise draws
$\boldsymbol\epsilon_1$ and $\boldsymbol\epsilon_2$
are shared across all models.
Let $\mathbf h_{m,a}^{\mathrm{ds}}$ denote the SAR tokens
after dual-stream block 8 for model $m$ and noise draw $a$.
We first compute the change in each model's SAR tokens:
\begin{equation}
\Delta\mathbf h_m^{\mathrm{ds}}
=\mathbf h_{m,2}^{\mathrm{ds}}-\mathbf h_{m,1}^{\mathrm{ds}}.
\label{eq:supp_eo_response}
\end{equation}
We then measure
$\cos(\Delta\mathbf h_{\mathcal T}^{\mathrm{ds}},
      \Delta\mathbf h_m^{\mathrm{ds}})$
for each student $m$, where $\mathcal T$ denotes the teacher.
A higher cosine means that changing the noisy EO input
moves the student's SAR tokens in a direction more
similar to the teacher's.

SPA achieves higher response cosine than the Baseline
at all four evaluated timesteps
$t\in\{0,0.25,0.50,0.75\}$, including
$0.658$ versus $0.355$ at $t=0.25$ and
$0.408$ versus $0.232$ at $t=0$.
Together with the lower MSE and higher CKA in panel~(a),
these results support the transfer of the teacher's
EO-conditioned SAR representations into SPA:
SPA better preserves both the representations themselves
and their responses to changes in noisy EO inputs.

\subsection{Quality of SAR Conditioning Features}
\label{supp:sar_readout}

\begin{figure}[!t]
\centering
\includegraphics[width=\columnwidth]{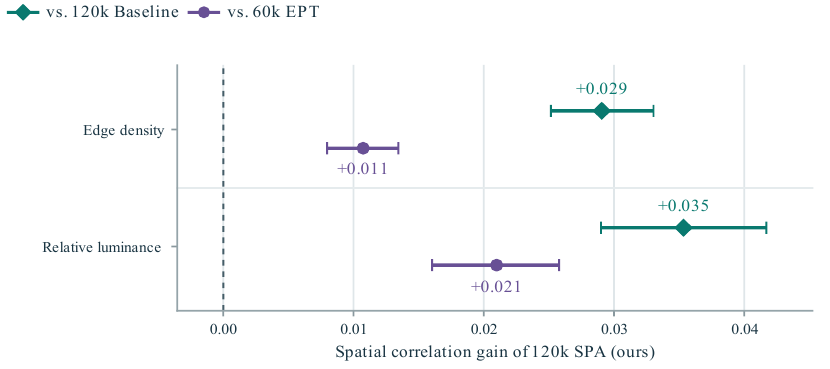}
\caption{\textbf{EO structural readout from frozen SAR features.}
Mean paired gains in within-tile Pearson correlation for
linear ridge prediction of GT EO edge density and relative
luminance from EMA SAR tokens at $t=0$.
The 120k SPA model is compared with the Baseline in
Table~\mainref{tab:phase_b_ablation} (b) and 60k EPT.
Bars indicate 95\% paired tile-bootstrap intervals.}
\label{fig:supp_sar_readout}
\end{figure}

We next assess the utility of the final SAR conditioning
features independently of teacher alignment.
Figure~\ref{fig:supp_sar_readout} compares 120k SPA with
the Baseline in Table~\mainref{tab:phase_b_ablation} (b) and
the original 60k EPT model with its VAE-based SAR pathway.
We extract frozen EMA tokens from SAR block 8 at $t=0$,
where the EO input is pure Gaussian noise.

We used 1,024 SAR--EO tile pairs from the QXS training split
to fit the probes and 1,024 from the test split for evaluation.
For each checkpoint, we fit a separate linear ridge probe
with a fixed penalty of $\lambda=0.01$ to predict GT EO
edge density and relative luminance at each token location.
We computed Pearson's $r$ across the $16\times16$ locations
within each test tile and averaged the paired differences
between checkpoints over tiles.

SPA improves the mean correlation over the Baseline by
$0.029$ and $0.035$, respectively, and over 60k EPT by
$0.011$ and $0.021$.
The 95\% paired tile-bootstrap intervals lie above zero
for all four comparisons.
These results indicate that SPA improves the linear
predictability of EO structural attributes over the
Baseline while retaining useful EO-related information
after removing the pretrained SAR VAE encoder.

\section{Pixel-wise Uncertainty Flow Matching}
\label{supp:ufm}

Prior work on SAR-to-EO translation~\cite{do2026c} begins
training with a standard diffusion noise-prediction loss
and later switches to a confidence-guided diffusion loss
based on $\beta$-NLL. Motivated by this schedule, we use
standard flow matching during EPT to adapt the EO generator
to pixel space and introduce PUFM at the start of SPA.

Below, we derive PUFM from a residual likelihood, describe
how we parameterize the variance, and analyze how $\beta$
affects the reconstruction gradient. We also compare
$\beta\in\{0,0.5,1\}$ under matched training settings.
Our choice of $\beta=1$ preserves the standard FM gradient
with respect to the predicted EO image while allowing the
model to learn pixel-wise residual variance.

\subsection{From Confidence-Guided Diffusion to PUFM}
C-DiffSET~\cite{do2026c} introduces confidence-guided
supervision to address imperfect SAR--EO correspondence
in latent diffusion.
Our PUFM shares this motivation but models the variance of
normalized flow residuals directly at the output-pixel
resolution. We use detached variance weighting motivated
by $\beta$-NLL~\cite{seitzer2022pitfalls}.
Pixel-resolution prediction permits spatially localized
uncertainty estimates without latent compression; this
architectural distinction does not by itself establish
more accurate uncertainty estimation.

To adapt uncertainty supervision to clean-image
flow prediction, we formulate a Gaussian likelihood
over pixel-space flow residuals.
For pixel $p$ and RGB channel $j$, let
$r_{pj}=(\hat x^e_{pj}-x^e_{pj})/d(t)$,
where $d(t)=\max(1-t,0.05)$.
The RGB-averaged squared FM residual is
$e_p=\frac13\sum_{j=1}^{3}r_{pj}^2$
$=\frac13\sum_{j=1}^{3}\ell_{\mathrm{FM},pj}$,
where $\ell_{\mathrm{FM},pj}$ is the elementwise loss
in Eq.~\mainref{eq:prelim_fm_loss}. Here $p$ indexes output pixels.
Assuming a Gaussian residual distribution,
\begin{equation}
 p(r_{pj}\mid q_{v,p})
 =\frac{1}{\sqrt{2\pi q_{v,p}}}
 \exp\left(-\frac{r_{pj}^2}{2q_{v,p}}\right),\\
 q_{v,p}>0,
 \label{eq:supp_residual_likelihood}
\end{equation}
where $q_{v,p}$ is the value of the main paper's variance
map $q_v$ at pixel $p$, shared across RGB channels.
Twice the RGB-averaged negative log-likelihood yields
$e_p/q_{v,p}+\log(2\pi q_{v,p})$.
Following the detached variance weighting of
$\beta$-NLL~\cite{seitzer2022pitfalls}, we define
\begin{equation}
\begin{aligned}
 u_{\beta,p}
 &=\sg[q_{v,p}]^\beta
   \left(\frac{e_p}{q_{v,p}}+\log(2\pi q_{v,p})\right),\\
 \mathcal L_{\mathrm{PUFM}}
 &=\mathbb E[u_{\beta,p}].
\end{aligned}
\label{eq:supp_ufm_element}
\end{equation}
The expectation averages over images, timesteps,
and spatial positions.
We omit the conventional $1/2$ factor to preserve
the scale of the direct FM reconstruction gradient
when $\beta=1$, ensuring consistency with the
standard FM objective used in EPT.
This variance models the normalized residual $r_{pj}$.
The corresponding clean-image residual variance is
$d(t)^2q_{v,p}$, so raw variance values across timesteps
are not directly comparable as EO intensity uncertainty.

\paragraph{Variance prediction.}
A training-only variance head predicts $q_{v,p}$ from the final RMS-normalized pixel-decoder feature $\mathbf f_p\in\mathbb R^{16}$:
\begin{equation}
\begin{aligned}
q_{v,p} &= \rho_v(\mathbf w_v^\top\mathbf f_p+b_v),\\
\rho_v(a) &= \operatorname{clip}_{[q_{\min},q_{\max}]}
\bigl(\operatorname{SSP}(a)\bigr),\\
\operatorname{SSP}(a) &= \log(1+(e-1)e^a),
\end{aligned}
\label{eq:supp_variance_prediction}
\end{equation}
where $\operatorname{SSP}$ denotes the shifted softplus,
$q_{\min}=10^{-6}$, and $q_{\max}=10.0$.
The shift ensures $\rho_v(0)=1$, while clipping
ensures numerical stability and prevents excessively
large variance estimates. The resulting variance map has the same spatial resolution as the generated EO image, with one variance value shared across RGB channels. Unlike uncertainty estimation on compressed latent features, this formulation directly associates uncertainty with pixel-space reconstruction errors, allowing finer spatial localization of potentially unreliable correspondences. The variance head receives gradients during SPA and LFA but is omitted during standard inference.

\subsection{Gradient Dependence on \texorpdfstring{$\beta$}{beta}}
\label{supp:beta_gradients}

The exponent $\beta$ controls how predicted variance
affects the direct reconstruction gradient.
Differentiating Eq.~\ref{eq:supp_ufm_element}, with the
outer variance factor detached and $\hat x^e_{pj}$ and
$q_{v,p}$ treated as independent output coordinates, gives
\begin{equation}
\begin{aligned}
\frac{\partial u_{\beta,p}}{\partial\hat x^e_{pj}}
&=q_{v,p}^{\beta-1}
  \frac{2(\hat x^e_{pj}-x^e_{pj})}{3d(t)^2},\\
\frac{\partial u_{\beta,p}}{\partial q_{v,p}}
&=q_{v,p}^{\beta-2}(q_{v,p}-e_p).
\end{aligned}
\label{eq:supp_ufm_gradients}
\end{equation}
The first expression is the standard FM gradient multiplied
by $q_{v,p}^{\beta-1}$.
Table~\ref{tab:supp_beta_gradients} summarizes the three
settings evaluated below.
At $\beta=0$, reconstruction uses inverse-variance weighting.
At $\beta=0.5$, inverse-standard-deviation weighting reduces
the dependence on predicted variance.
Both settings assign relatively smaller reconstruction
weights to pixels with larger predicted variance.
At $\beta=1$, the variance factor cancels, leaving the
direct FM reconstruction gradient independent of $q_{v,p}$.

All three settings retain variance supervision through
the second expression in Eq.~\ref{eq:supp_ufm_gradients}.
For a fixed positive $e_p$, the variance-coordinate
stationary point is $q_{v,p}=e_p$ in the unclipped regime.
At the clipping bounds, the gradient to the raw variance
logit can vanish, so this stationary-point interpretation
applies only to the unclipped variance coordinate.

\begin{table}[!t]
\centering\GeoSuppTableStyle
\caption{\textbf{Effect of $\beta$ on the direct FM gradient.}
The multiplier scales the standard FM reconstruction
gradient in Eq.~\ref{eq:supp_ufm_gradients}.
All three settings retain variance supervision.}
\label{tab:supp_beta_gradients}
\begin{tabularx}{\columnwidth}
{cc>{\raggedright\arraybackslash}X}
\toprule
$\beta$ & FM multiplier & Interpretation\\
\midrule
$0$ & $q_{v,p}^{-1}$ & Inverse-variance weighting\\
$0.5$ & $q_{v,p}^{-1/2}$ & Inverse-standard-deviation weighting\\
\GeoSuppOurRow
$1$ & $1$ & Variance-independent direct FM gradient\\
\bottomrule
\end{tabularx}
\end{table}

\subsection{Sensitivity to the Uncertainty Exponent}
\label{supp:beta_ablation}

We compare $\beta\in\{0,0.5,1\}$ under matched settings.
All variants start from the same EPT checkpoint and undergo
60k SPA updates with identical joint optimization, CPD,
Gamma augmentation, curriculum, and inference settings.
Evaluation uses four Euler steps and six NFEs, before LFA
introduces fixed input times and call-dependent objectives.
We compare image quality rather than raw PUFM loss values,
whose scale and variance contribution depend on $\beta$.

Table~\ref{tab:supp_beta_ablation} shows that $\beta=1$
achieves the lowest FID of $16.8$, compared with $17.5$
for $\beta=0.5$ and $21.6$ for $\beta=0$.
The setting $\beta=0.5$ achieves the lowest DISTS and
LPIPS, whereas $\beta=0$ yields the highest SSIM and PSNR.
The empirical advantage of $\beta=1$ is therefore its
lower FID, with a trade-off in paired perceptual and
reconstruction metrics.

\begin{table}[!t]
\centering\GeoSuppTableStyle
\caption{\textbf{PUFM exponent ablation on QXS-SAROPT.}
All variants use the same SPA initialization, 60k-update
budget, and four-step evaluation (six NFEs).
$\beta=1$ achieves the lowest FID, while smaller exponents
favor SSIM and PSNR.}
\label{tab:supp_beta_ablation}
\begin{tabularx}{\columnwidth}
{c*{5}{>{\centering\arraybackslash}X}}
\toprule
$\beta$ & FID$\downarrow$ & DISTS$\downarrow$ & LPIPS$\downarrow$
& SSIM$\uparrow$ & PSNR$\uparrow$\\
\midrule
$0$ & 21.6 & .227 & .507
& \textbf{.402} & \textbf{17.50}\\
$0.5$ & 17.5 & \textbf{.219} & \textbf{.506} & .380 & 17.33\\
\GeoSuppOurRow
$1$ & \textbf{16.8} & .221 & .509 & .362 & 17.00\\
\bottomrule
\vspace{-7px}
\end{tabularx}
\end{table}

To interpret this empirically favored setting, substituting
$\beta=1$ into Eq.~\ref{eq:supp_ufm_gradients} gives
\begin{equation}
\begin{aligned}
\frac{\partial u_{1,p}}{\partial\hat x^e_{pj}}
&=\frac{2(\hat x^e_{pj}-x^e_{pj})}{3d(t)^2},\\
\frac{\partial u_{1,p}}{\partial q_{v,p}}
&=1-\frac{e_p}{q_{v,p}}.
\end{aligned}
\label{eq:supp_beta1_gradients}
\end{equation}
The direct clean-EO reconstruction gradient is identical
to standard FM, while the variance branch continues to
learn pixel-wise residual uncertainty.
For parameters $\vartheta\subseteq\Theta$ shared by the
RGB and variance branches, the gradient becomes
\begin{equation}
\nabla_\vartheta u_{1,p}
=\nabla_\vartheta e_p
+\left(1-\frac{e_p}{q_{v,p}}\right)\nabla_\vartheta q_{v,p}.
\label{eq:supp_beta1_shared_gradient}
\end{equation}
Thus, at $\beta=1$, PUFM combines standard FM reconstruction
supervision with auxiliary estimation of pixel-wise
normalized residual variance.
The variance-learning term can influence shared generator
representations through $\nabla_\vartheta q_{v,p}$, while
the direct reconstruction gradient remains independent
of predicted variance.
In particular, larger predicted variance does not reduce
a pixel's direct FM reconstruction gradient.

We therefore adopt $\beta=1$ for its lowest observed FID
among the evaluated settings and its ability to retain
uncertainty supervision without variance-dependent
reweighting of the direct reconstruction gradient.
The learned variance characterizes normalized prediction
residuals, which may reflect modality ambiguity,
misregistration, or prediction error; it does not
distinguish these causes separately.

\subsection{Pixel-Space Uncertainty Maps}
\label{supp:uncertainty_qualitative}

\begin{figure}[!t]
\centering
\includegraphics[width=\columnwidth]
{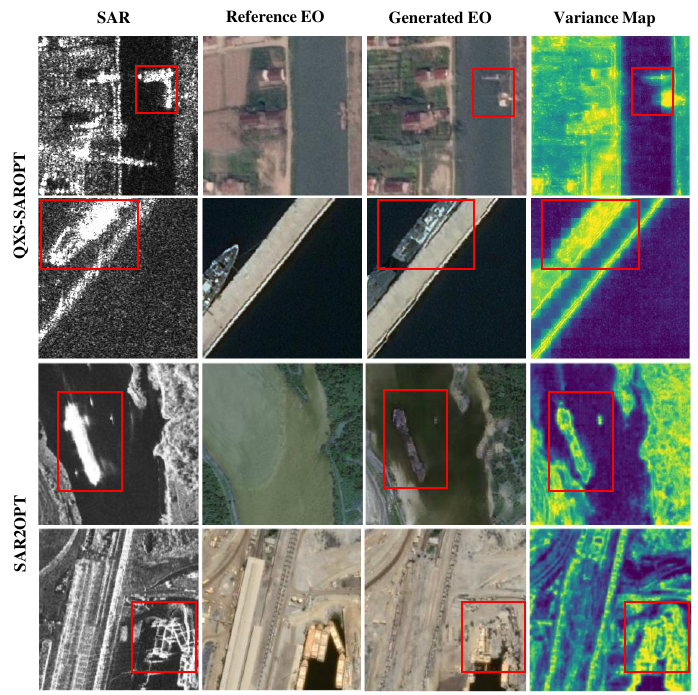}
\caption{\textbf{Pixel-space uncertainty in two-step LFA.}
Columns show SAR, reference EO, generated EO, and the variance map.
Each variance map is independently min--max normalized to $[0,1]$
for visualization, with dark purple and yellow indicating values
near $0$ and $1$, respectively.
The first two rows are from QXS-SAROPT and the last two from SAR2Opt.}
\label{fig:supp_pufm_uncertainty}
\vspace{-7px}
\end{figure}

Figure~\ref{fig:supp_pufm_uncertainty} shows the second
clean-EO prediction $\hat{\mathbf x}^{e,(2,2)}$ at $t=0.5$
and the predicted pixel-wise variance $q_v$ of its
normalized flow residual. For this visualization, we
retain the variance head to read out $q_v$ during sampling.
As discussed above, PUFM with $\beta=1$ preserves the direct
FM reconstruction gradient while learning pixel-wise
residual uncertainty (Eq.~\ref{eq:supp_beta1_gradients}).
In the red-boxed regions, structural discrepancies between
SAR and the reference EO coincide with elevated predicted
variance, illustrating how PUFM represents uncertainty
associated with imperfect SAR--EO correspondence.

\section{Training Implementation and Dynamics}
\label{supp:training}
\label{supp:losses}

\subsection{SAR Augmentation}
\label{supp:gamma}

Gamma speckle augmentation is applied during SPA and LFA training.
After paired geometric transformations, let
$\mathbf a^s=(\mathbf x^s+1)/2$ denote the SAR image
mapped to $[0,1]$.
For a retained SAR selected for augmentation,
we use
\begin{equation}
\tilde{\mathbf x}^s
=2\operatorname{clip}_{[0,1]}
\left(
\mathbf a^s+\kappa
\mathbf a^s\odot(\mathbf G-1)
\right)-1,
\label{eq:supp_gamma}
\end{equation}
where $\mathbf G=(G_{ij})$ and
$G_{ij}\mid L\sim\operatorname{Gamma}(L,L)$ in the
shape--rate parameterization. Thus,
$\mathbb E[G_{ij}\mid L]=1$ and
$\operatorname{Var}(G_{ij}\mid L)=1/L$.
The values of $L$, $\kappa$, and the augmentation
probability follow Sec.~\mainref{sec:exp_setup}.
Noise is sampled independently at each spatial location
and shared across replicated SAR channels.
The frozen teachers and inference use the original SAR
observation.
We write $\tilde{\mathbf x}^s=\mathbf x^s$ when
augmentation is not applied.

\paragraph{Effect of speckle augmentation during SPA.}
The student sees perturbed SAR intensities on a subset of training
examples while the EPT teacher supplies features from the original
SAR observation. To evaluate whether this perturbation improves
translation on the original test inputs, Table~\ref{tab:supp_spa_augmentation} compares the full SPA setting
with a control that sets the student's Gamma augmentation probability
$p_\Gamma$ to zero. Here $p_\Gamma$ is conditional on retaining the
SAR input. The comparison keeps paired geometric augmentation and
the $0.1$ SAR conditioning dropout probability. Student and teacher
share the dropout mask, and the teacher receives the unperturbed
SAR observation whenever the condition is retained.

The reference protocol initializes the student from the raw 60k EPT
checkpoint and uses its frozen EMA copy for CPD. The matched control
calls for the same 60k SPA updates with joint optimization, CPD, and
PUFM, followed by evaluation on QXS-SAROPT with four Euler steps
and the guidance schedule in Sec.~\mainref{sec:exp_setup}.

\begin{table}[!t]
\centering\GeoSuppTableStyle
\caption{\textbf{Effect of training-time Gamma speckle augmentation
during SPA on QXS-SAROPT.}
The $p_\Gamma=0$ row disables Gamma speckle augmentation only,
while the $p_\Gamma=0.5$ row corresponds to full SPA in
Table~\mainref{tab:phase_b_ablation}(b).}
\label{tab:supp_spa_augmentation}
\setlength{\tabcolsep}{2.1pt}
\begin{tabular}{@{}crrrrr@{}}
\toprule
$p_\Gamma$ & FID$\downarrow$ & DISTS$\downarrow$ & LPIPS$\downarrow$
& SSIM$\uparrow$ & PSNR$\uparrow$\\
\midrule
$0$ & 17.0 & \textbf{.221} & .512 & .361 & \textbf{17.03} \\
$0.5$ & \textbf{16.8} & \textbf{.221} & \textbf{.509} & \textbf{.362} & 17.00\\
\bottomrule
\end{tabular}
\end{table}

During SPA, we apply Gamma speckle augmentation to the SAR input
with probability $p_\Gamma=0.5$. On QXS-SAROPT, enabling
augmentation slightly improves FID (17.0 to 16.8), LPIPS
(0.512 to 0.509), and SSIM (0.361 to 0.362). DISTS is 0.221
for both settings at the reported precision, while PSNR decreases
from 17.03 to 17.00\,dB.
The small differences may reflect the limited number of training
pairs and the limited diversity of satellite observations in
QXS-SAROPT. We hypothesize that Gamma speckle augmentation
could yield larger gains when training on a larger dataset
that includes observations from multiple SAR satellites
and covers a broader range of speckle statistics.

\subsection{Satellite Feature Extraction}
\label{supp:feature_networks}
We use the frozen DINOv3-SAT \texttt{ViT-L/16-SAT493M} encoder. We rescale inputs as $(\mathbf u+1)/2$ and normalize them
using channel means $(0.430,0.411,0.296)$ and standard
deviations $(0.213,0.156,0.143)$.
Before rescaling, we clip REPA targets to $[-1,1]$, use
the same bounds with a straight-through gradient for generated
EO in the adversarial loss, and leave both inputs to
$\ell_{\mathrm{SAT}}$ unclipped. For REPA in Eq.~\mainref{eq:repa}, EO tokens from the final dual-stream DiT block ($L=8$) are mapped to the last-layer satellite features using a three-layer, width-1024 linear projector with SiLU activations between layers. For $\ell_{\mathrm{SAT}}$ in Eq.~\mainref{eq:sat_perceptual}, we use the one-indexed layers $\mathcal K=\{2,5,9,13,17,21\}$. Their patch features are processed by the encoder's common final LayerNorm, excluding CLS and register tokens. Target-side features are detached, whereas the frozen feature extractors still propagate gradients from generated-EO features to the generator.

\subsection{EO-Only Adversarial Training}
\label{supp:adversarial}
The discriminator provides EO appearance supervision, complementing the paired losses and SAR conditioning used by the generator. For each $k\in\mathcal K$, the trainable head $h_{\psi_k}$ consists of LayerNorm$_{1024}$, spectral-normalized Linear$_{1024\to512}$, LeakyReLU$(0.2)$, and spectral-normalized Linear$_{512\to1}$. As in the main paper, $C_{\psi_k}=h_{\psi_k}\circ\mathcal F_\phi^{(k)}$ includes the frozen feature extractor:
\begin{equation}
\begin{aligned}
\ell_D(\mathbf u,\mathbf v)
={}&\sum_{k\in\mathcal K}\bigl[
\ell_{\mathrm{BCE}}(C_{\psi_k}(\mathbf v),1)\\
&+\ell_{\mathrm{BCE}}(C_{\psi_k}(\mathbf u),0)
\bigr],\\
\mathcal L_D
={}&\lambda_D\mathbb E_{\mathcal B}
\bigl[
\ell_D(\sg[\hat{\mathbf x}^e],\mathbf x^e)
\bigr].
\end{aligned}
\label{eq:supp_discriminator_raw}
\end{equation}
Here $\ell_{\mathrm{BCE}}$ already averages over patch locations, consistently with Eq.~\mainref{eq:gan_objective}. The generator's adversarial target remains $1$.

 For adversarial feature extraction, we clip generated EO
images to $[-1,1]$ before rescaling and normalization.
During generator updates, the clip uses a straight-through
estimator with an identity gradient. The DINOv3-SAT encoder remains frozen, discriminator parameters are fixed during generator updates, and generated images are detached during discriminator updates.

\begin{table*}[!t]
\centering
\footnotesize
\caption{\textbf{Base loss coefficients by phase,
curriculum stage, and student call.}
Numerical entries are reported before the applicable
timestep weighting and thresholding.
LP/SAT denotes LPIPS/satellite-perceptual losses,
and GAN lists generator/discriminator coefficients.
T-MAE denotes MAE supervision against the frozen SPA
teacher endpoint.
T-LP/SAT denotes LPIPS/satellite-perceptual supervision
against the same endpoint, with base coefficients
$\lambda_{\mathcal T,P}\lambda_{\mathrm{LP}}$ and
$\lambda_{\mathcal T,P}\lambda_{\mathrm{SAT}}$, respectively.
LFA weighting is specified in Sec.~\ref{supp:lfa_weights}.}
\label{tab:training_schedule}
\label{tab:sschedule}
\setlength{\tabcolsep}{4pt}
\renewcommand{\arraystretch}{1.16}

\begin{tabularx}{\textwidth}
{@{}lcl*{6}{>{\centering\arraybackslash}X}@{}}
\toprule
\makecell[l]{Phase}
& \makecell[c]{Stage /\\call}
& \makecell[c]{Flow}
& \makecell[c]{REPA\\
  $\lambda_{\mathrm{REPA}}$}
& \makecell[c]{Paired\\LP / SAT\\
  $\lambda_{\mathrm{LP}}\,/\,\lambda_{\mathrm{SAT}}$}
& \makecell[c]{GAN (G / D)\\
  $\lambda_{\mathrm{GAN}}\,/\,\lambda_D$}
& \makecell[c]{CPD\\
  $\lambda_{\mathrm{CPD}}$}
& \makecell[c]{T-MAE\\
  $\lambda_{\mathcal T,1}^{\mathrm{base}}$}
& \makecell[c]{T-LP / SAT\\
  $\lambda_{\mathcal T,P}\lambda_{\mathrm{LP}}$\\
  $/\,\lambda_{\mathcal T,P}\lambda_{\mathrm{SAT}}$}\\
\midrule

EPT & 1 & FM
& 0.5 & -- & -- & -- & -- & --\\
EPT & 2 & FM
& 0.5 & 0.1 / 0.01 & -- & -- & -- & --\\
EPT & 3 & FM
& 0.5 & 0.1 / 0.01 & 0.05 / 1 & -- & -- & --\\

\midrule

SPA & 1 & PUFM
& 0.5 & -- & -- & 1 & -- & --\\
SPA & 2 & PUFM
& 0.5 & 0.1 / 0.01 & -- & 1 & -- & --\\
SPA & 3 & PUFM
& 0.5 & 0.1 / 0.01 & 0.05 / 1 & 1 & -- & --\\

\midrule

LFA ($n=1$) & $r=1$ & PUFM
& 0.25 & 1 / 1 & 0.6 / 0.6
& -- & 1.0 & 0.5 / 0.5\\

LFA ($n=2$) & $r=1,2$ & PUFM
& 0.25 & 0.5 / 0.5 & 0.3 / 0.6
& -- & 0.25 & 0.25 / 0.25\\

\bottomrule
\vspace{-7px}
\end{tabularx}
\end{table*}

\subsection{Training Algorithms}
\label{supp:end_to_end}

Algorithms~\ref{alg:whole} and~\ref{alg:lfa} summarize
EPT--SPA training and independent LFA training, respectively,
using the objectives defined in the main paper.
Algorithm-specific notation is defined within the pseudocode.

\noindent\textbf{Pseudocode conventions.}
$\operatorname{Update}(\Omega,\mathcal L)$ denotes an AdamW
step on $\Omega$ with gradient clipping at norm $1$,
using the settings in Sec.~\mainref{sec:exp_setup}.
We write
$\operatorname{EMA}(\bar\Theta,\Theta)
=0.999\bar\Theta+0.001\Theta$.
$\operatorname{Aug}_\Gamma$ applies the Gamma augmentation
in Eq.~\ref{eq:supp_gamma} with probability $0.5$ per example
and otherwise returns the input unchanged.
$\operatorname{Cond}_c$ returns the input for $c=1$ and the
null SAR condition for $c=0$.
$\ell_D$ follows Eq.~\ref{eq:supp_discriminator_raw},
and $\mathbb E_{\mathcal B}$ denotes minibatch averaging.
Table~\ref{tab:training_schedule} gives the base loss coefficients.

\begin{algorithm}[!t]
\caption{\textbf{Pixel-space conversion: EPT $\to$ SPA}}
\label{alg:whole}
\footnotesize
\algrenewcommand\algorithmicindent{0.8em}
\begin{algorithmic}[1]

\Require Dataset $\mathcal D$, pretrained ReFlowSET parameters
$\Theta_0$, REPA projector $\omega$, and discriminator $\Psi$.
\Ensure Phase checkpoints
$(\Theta_P,\bar\Theta_P,\omega_P,\Psi_P)$
for $P\in\{\mathrm{EPT},\mathrm{SPA}\}$.

\Statex $p_1=\operatorname{LogitNormal}(-0.8,0.8^2)$;
$p_2=p_3=\mathcal U(0,1)$.

\State $\Theta\gets\Theta_0$; replace the EO input interface
and decoder with EO-LPA and PiT.

\For{$P\in(\mathrm{EPT},\mathrm{SPA})$}
    \If{$P=\mathrm{SPA}$}
        \State $\Theta\gets\Theta_{\mathrm{EPT}}$;
        freeze
        $\mathcal T_{\mathrm{EPT}}
        \gets\mathcal G_{\bar\Theta_{\mathrm{EPT}}}^{\mathrm{EPT}}$.
        \State Replace the SAR input interface with a copy
        of the EPT-trained EO-LPA.
    \EndIf

    \State $\Omega^P\gets\Theta\cup\{\omega\}$;
    remove $\Theta_S$ if $P=\mathrm{EPT}$.

    \For{$j=1,\ldots,60000$}
        \State $i\gets1+\lfloor(j-1)/20000\rfloor$
        \State Sample $\mathcal B\sim\mathcal D$,
        $\boldsymbol\epsilon\sim\mathcal N(\mathbf0,\mathbf I)$,
        $t\sim p_i$, and $c\sim\operatorname{Bern}(0.9)$
        per example.

        \State $\tilde{\mathbf x}^s\gets\mathbf x^s$;
        if $P=\mathrm{SPA}$, set
        $\tilde{\mathbf x}^s\gets
        \operatorname{Aug}_\Gamma(\mathbf x^s)$.
        \State $\mathbf x_t^e
        \gets(1-t)\boldsymbol\epsilon+t\mathbf x^e$
        \State $\hat{\mathbf x}^e
        \gets\mathcal G_\Theta^P
        (\mathbf x_t^e,t,
        \operatorname{Cond}_c(\tilde{\mathbf x}^s))$

        \State $\mathcal L_G
        \gets\mathcal L_{\mathrm{curr}}^P(i)$
        \State If $P=\mathrm{SPA}$, add
        $\lambda_{\mathrm{CPD}}\mathcal L_{\mathrm{CPD}}$
        to $\mathcal L_G$.
        \Statex $\triangleright$ CPD teacher input:
        $(\mathbf x_t^e,t,
        \operatorname{Cond}_c(\mathbf x^s))$.

        \State If $j=40001$, set $\bar\Theta\gets\Theta$.
        \State $\Omega^P\gets
        \operatorname{Update}(\Omega^P,\mathcal L_G)$

        \If{$i=3$}
            \State $\mathcal L_D\gets
            \lambda_D\mathbb E_{\mathcal B}
            [\ell_D(\sg[\hat{\mathbf x}^e],\mathbf x^e)]$
            \State $\Psi\gets
            \operatorname{Update}(\Psi,\mathcal L_D)$
            \State $\bar\Theta\gets
            \operatorname{EMA}(\bar\Theta,\Theta)$
        \EndIf
    \EndFor

    \State Save $(\Theta_P,\bar\Theta_P,\omega_P,\Psi_P)
    \gets(\Theta,\bar\Theta,\omega,\Psi)$.
\EndFor

\end{algorithmic}
\end{algorithm}

\begin{algorithm}[!t]
\caption{\textbf{Independent low-step adaptation:
$n\in\{1,2\}$}}
\label{alg:lfa}
\footnotesize
\algrenewcommand\algorithmicindent{0.8em}
\begin{algorithmic}[1]

\Require Dataset $\mathcal D$ and SPA checkpoints
$(\Theta_{\mathrm{SPA}},\bar\Theta_{\mathrm{SPA}},
\omega_{\mathrm{SPA}},\Psi_{\mathrm{SPA}})$.
\Ensure Independently trained one- and two-step generators
and their EMA copies.

\Statex $\hat{\mathbf x}_r$ denotes prediction $r$ of student $n$.
$\sg_\eta$ follows Eq.~\ref{eq:supp_gradient_scaling}.

\State Freeze
$\mathcal T\gets
\mathcal G_{\bar\Theta_{\mathrm{SPA}}}^{\mathrm{SPA}}$;
set CFG scales $\mathbf s=(1,2.5,2.5,1)$.

\For{$n\in\{1,2\}$}
    \State $(\Theta^{(n)},\omega^{(n)},\Psi^{(n)})\gets$
    $(\Theta_{\mathrm{SPA}},\omega_{\mathrm{SPA}},
    \Psi_{\mathrm{SPA}})$
    \State $\bar\Theta^{(n)}\gets\Theta^{(n)}$;
    $\Omega^{(n)}\gets\Theta^{(n)}\cup\{\omega^{(n)}\}$

    \For{$j=1,\ldots,10000$}
        \State Sample $\mathcal B\sim\mathcal D$,
        $\boldsymbol\epsilon\sim\mathcal N(\mathbf0,\mathbf I)$,
        and $c\sim\operatorname{Bern}(0.9)$ per example.

        \State $\bar{\mathbf x}_{\mathcal T}
        \gets\operatorname{Euler}_{4,\mathbf s}
        (\mathcal T;\boldsymbol\epsilon,\mathbf x^s)$
        \State $\bar{\mathbf x}_{\mathcal T}
        \gets\sg[
        \operatorname{clip}_{[-1,1]}
        (\bar{\mathbf x}_{\mathcal T})]$

        \State $\tilde{\mathbf x}^s
        \gets\operatorname{Aug}_\Gamma(\mathbf x^s)$
        \State $\mathbf u_0\gets\boldsymbol\epsilon$;
        $\mathcal L_G\gets0$; $\mathcal L_D\gets0$

        \For{$r=1,\ldots,n$}
            \State $t\gets(r-1)/n$
           \State $\hat{\mathbf x}_r
            \gets\mathcal G_{\Theta^{(n)}}^{\mathrm{LFA}}
            (\mathbf u_{r-1},t,
            \operatorname{Cond}_c(\tilde{\mathbf x}^s))$

            \State $\mathcal L_G\gets\mathcal L_G
            +\mathcal L_{\mathrm{curr}}^{(n,r)}
            +\mathcal L_{\mathcal T}^{(n,r)}$
            \State $\mathcal L_D\gets\mathcal L_D
            +\lambda_D
            \mathbb E_{\mathcal B}
            [\ell_D(\sg[\hat{\mathbf x}_r],\mathbf x^e)]$

            \If{$r<n$}
                \State $\mathbf u_1\gets
                \frac12\bigl(
                \boldsymbol\epsilon+
                \sg_{0.1}(\hat{\mathbf x}_1)
                \bigr)$
            \EndIf
        \EndFor

        \State $\Omega^{(n)}\gets
        \operatorname{Update}(\Omega^{(n)},\mathcal L_G)$
        \State $\Psi^{(n)}\gets
        \operatorname{Update}(\Psi^{(n)},\mathcal L_D)$
        \State $\bar\Theta^{(n)}\gets
        \operatorname{EMA}(\bar\Theta^{(n)},\Theta^{(n)})$
    \EndFor
\EndFor

\end{algorithmic}
\end{algorithm}

\subsection{Training Dynamics and Test-Set Performance}
\label{supp:training_dynamics}

\paragraph{Training losses.}
Figure~\ref{fig:supp_training_curves} shows the total
training losses during EPT, SPA, and the independently
trained one- and two-step LFA continuations.
The right panel enlarges the LFA interval, while dashed
vertical lines mark the phase transitions at 60k and
120k updates.
Since the objectives and loss coefficients differ across
phases, the loss magnitudes should be interpreted within
each phase.

\paragraph{Test-set performance.}
Figure~\ref{fig:supp_validation_curves} tracks translation
quality on the QXS-SAROPT test set throughout training. The EPT curve shows the full three-stage run.
Each curriculum variant in Table~\mainref{tab:phase_a_ablation}(b)
received 60k updates, divided equally among its stages.
EPT and SPA show overall improvements in FID and DISTS
as training progresses.
At the transition to LFA, evaluation switches from
four-step sampling with CFG to one- or two-step sampling
without CFG; the discontinuity at 120k therefore also
reflects this change in inference configuration.
Within each fixed low-step setting, subsequent LFA updates
improve FID, DISTS, and LPIPS.
The accompanying changes in PSNR and SSIM illustrate
the trade-off between perceptual quality and pixel fidelity.

\begin{figure*}[!t]
\centering
\SuppAsset{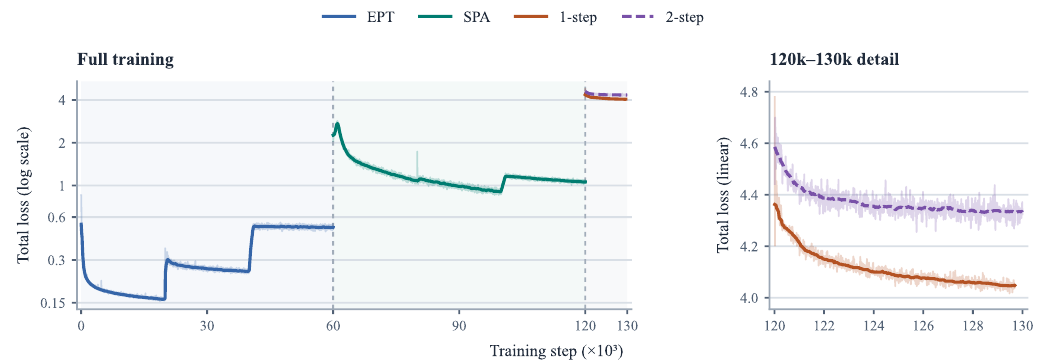}{4.2cm}{%
Training curves awaiting logs\\[5pt]
EPT \quad SPA \quad LFA 1-step \quad LFA 2-step\\[3pt]
Show both two-step prediction losses and curriculum transitions.}
\caption{\textbf{Training dynamics of EPT, SPA, and LFA.}
Left: total training losses during EPT (0--60k), SPA
(60--120k), and the independently trained one- and two-step
LFA continuations (120--130k).
Right: an enlarged view of the LFA interval.
Dashed vertical lines mark the EPT-to-SPA and SPA-to-LFA
phase boundaries.
Loss magnitudes depend on the phase-specific objectives
and coefficients.}
\label{fig:supp_training_curves}
\end{figure*}

\begin{figure*}[!t]
\centering
\includegraphics[width=\textwidth]
{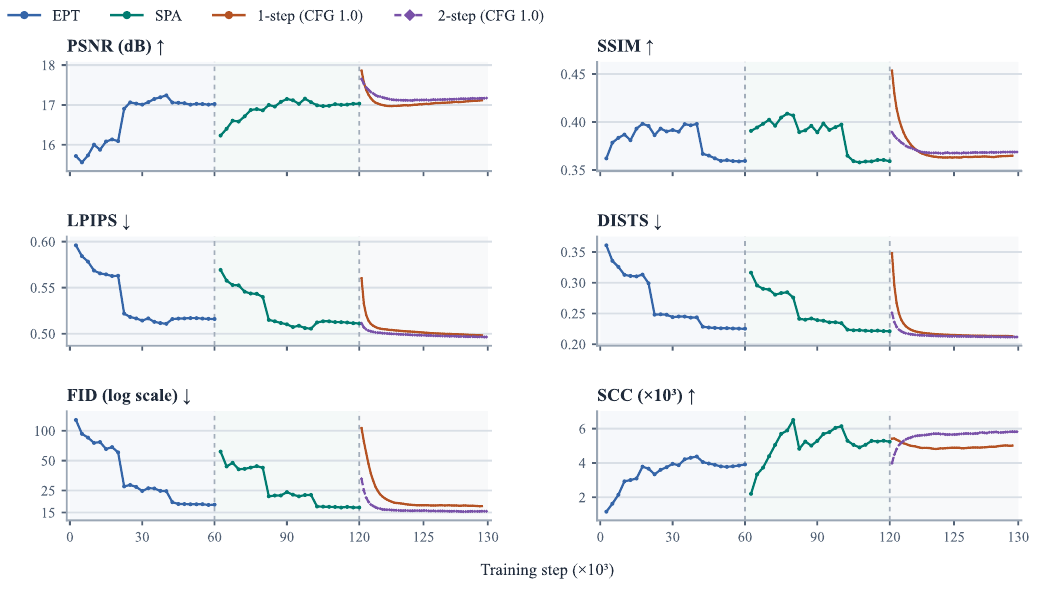}
\caption{\textbf{Test-set performance during EPT, SPA, and LFA
on QXS-SAROPT.} SCC denotes the spatial correlation coefficient between generated
and reference EO images and is scaled by $10^3$ for display.
The horizontal axis shows cumulative training updates,
with the 120k--130k interval expanded horizontally for visibility.
The one- and two-step LFA branches independently start
from the same 120k-update SPA checkpoint.
EPT and SPA use four Euler steps with CFG (six NFEs),
whereas the one- and two-step LFA models use one and two
NFEs without CFG, respectively.}
\label{fig:supp_validation_curves}
\vspace{-7px}
\end{figure*}

\section{Low-Step Inference and Larger-Patch Adaptation}
\label{supp:twostep}
\label{supp:lfa}

In our QXS-SAROPT experiments, the teacher achieves its
best observed performance with four-step sampling,
motivating our focus on one- and two-step LFA students.
More generally, if a teacher achieves its best performance
at a larger sampling budget, the same LFA formulation
could be extended to three-, four-, or other low-step
students by adjusting the rollout timesteps and
per-call loss coefficients.
We investigate the one- and two-step settings in this work.

\subsection{Inference-Step Sensitivity of SPA}
\label{supp:spa_step_sweep}

We examine how the inference budget affects the fully
trained SPA generator before LFA.
Figure~\ref{fig:supp_spa_step_sweep} evaluates the same
SPA checkpoint with one to ten Euler steps on QXS-SAROPT,
without additional training.
Among the evaluated budgets, four-step sampling achieves
the lowest FID, DISTS, and LPIPS.
Increasing the number of steps beyond four increases
latency without improving these metrics.

Direct one- and two-step sampling substantially worsens
the perceptual metrics relative to four-step sampling.
Although one-step sampling yields higher PSNR and SSIM,
reducing the inference budget alone does not preserve
the perceptual quality of the SPA generator.
These observations support our use of a four-step SPA
teacher and motivate dedicated adaptation to one- and
two-step inference through LFA.

\begin{figure*}[t]
\centering
\includegraphics[width=\textwidth]
{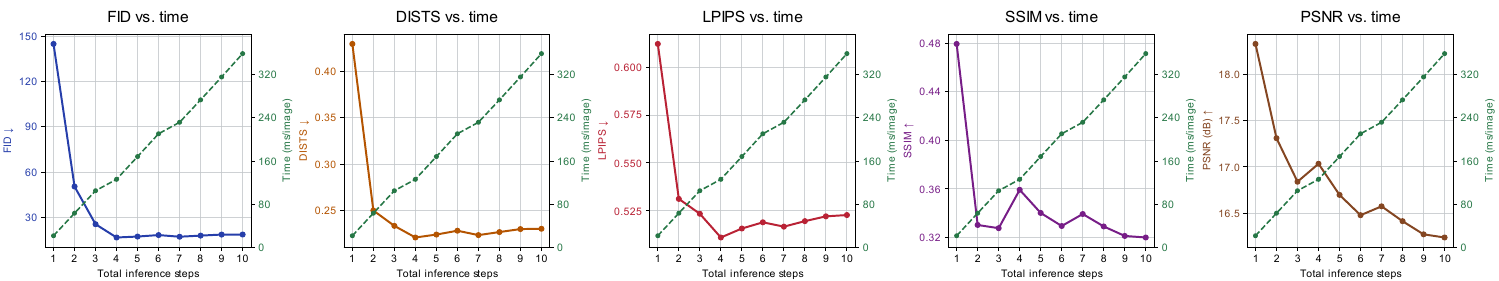}
\caption{
\textbf{Inference-step sensitivity of SPA on QXS-SAROPT.}
We evaluate the 120k-update SPA EMA checkpoint on
3,999 test pairs at $256\times256$ resolution using
one to ten Euler steps, without LFA.
Solid curves show the quality metrics on the left axes;
green dashed curves show mean latency on the right axes,
measured on one RTX A6000 with batch size one.
The CFG scale is $2.5$ for $t\in[0.1,0.7]$ and $1$ otherwise.
The four-step configuration uses six NFEs and achieves
the lowest FID, DISTS, and LPIPS among the evaluated budgets.
}
\label{fig:supp_spa_step_sweep}
\vspace{-10px}
\end{figure*}

\subsection{Students, Teacher, and Gradient Scaling}
\label{supp:rollout}
We use the students $\mathcal G_{\Theta^{(n)}}^{\mathrm{LFA}}$ and teacher $\mathcal T_{\mathrm{SPA}}$ defined in Sec.~\mainref{sec:lfa}. The teacher endpoint $\bar{\mathbf x}_{\mathcal T}$ is clipped to $[-1,1]$ and detached before supervision. Teacher sampling uses the same initial noise as the student and the SPA inference configuration in Sec.~\mainref{sec:exp_setup}. Paired EO images remain the targets for PUFM and REPA; the EPT-to-SPA CPD loss is not applied during LFA.

\paragraph{Gradient scaling.}
The connected two-step trajectory allows losses on the second prediction to backpropagate through the intermediate Euler state. To control the magnitude of this additional backward path, we use
\begin{equation}
\sg_\eta(\mathbf a)
=\sg[\mathbf a]+\eta(\mathbf a-\sg[\mathbf a]).
\label{eq:supp_gradient_scaling}
\end{equation}
This operation is the identity in the forward pass and scales its input gradient by $\eta$. It is applied only when forming an intermediate state; direct supervision on each clean-image prediction is unchanged.

With $\mathbf u_0=\boldsymbol\epsilon$, the first two-step prediction is
$\hat{\mathbf x}^{e,(2,1)}=\mathcal G_{\Theta^{(2)}}^{\mathrm{LFA}}(\mathbf u_0,0,\tilde{\mathbf x}^s)$.
The midpoint at $t_1=1/2$ is
\begin{equation}
\mathbf u_1=
\frac{1}{2}\left(
\boldsymbol\epsilon+
\sg_{0.1}(\hat{\mathbf x}^{e,(2,1)})
\right).
\label{eq:supp_midpoint}
\end{equation}
The second prediction is $\hat{\mathbf x}^{e,(2,2)}=\mathcal G_{\Theta^{(2)}}^{\mathrm{LFA}}(\mathbf u_1,1/2,\tilde{\mathbf x}^s)$, with no noise resampling. The midpoint Jacobian with respect to the first prediction is $0.05\mathbf I$, compared with $0.5\mathbf I$ for the unscaled Euler update. Thus, this backward path is attenuated by a factor of $0.1$ while remaining connected.

\subsection{Loss Weighting in LFA}
\label{supp:lfa_ufm}
\label{supp:lfa_weights}

Table~\ref{tab:training_schedule} reports base coefficients
before the applicable timestep weighting and thresholding.
For LFA, the appearance-loss gate is active at $t=0$ for the
one-step student. For the two-step student, it is inactive
at the first call ($t=0$) and active at the second
($t=0.5$). This rule applies to the paired perceptual,
generator-adversarial, and teacher perceptual terms in
Eq.~\mainref{eq:lfa_objective}.
The two-step student uses the same base coefficients at
both calls.

For teacher MAE, the coefficient $\lambda_{\mathcal T,1}$
in Eq.~\mainref{eq:lfa_objective} is the effective weight
$\lambda_{\mathcal T,1}^{\mathrm{base}}d(t_{r-1})^{-2}$.
We use this factor to give the teacher MAE the same timestep
weighting as the squared clean-image error in the flow-matching
loss in Eq.~\mainref{eq:prelim_fm_loss}.
Thus, the two-step base coefficient of $0.25$ gives effective
weights of $0.25$ at $t_0=0$ and $1$ at $t_1=0.5$.
The one-step weight remains $1$.
The effective LFA generator-loss coefficients derived from
Table~\ref{tab:training_schedule} are summarized below for
each student call $(n,r)$.
\begin{equation}
\label{eq:supp_lfa_effective_weights}
\setlength{\arraycolsep}{3pt}
\begin{array}{@{}l@{\quad}ccc@{}}
\toprule
\text{Loss}              & (1,1) & (2,1) & (2,2) \\
\midrule
\text{PUFM}              & 1    & 1    & 1    \\
\text{REPA}              & 0.25 & 0.25 & 0.25 \\
\text{Paired LPIPS/SAT}  & 1    & 0    & 1    \\
\text{Adversarial}       & 0.6  & 0    & 0.6  \\
\text{Teacher MAE}       & 1    & 0.25 & 1    \\
\text{Teacher LPIPS/SAT} & 0.5  & 0    & 0.5  \\
\bottomrule
\end{array}
\end{equation}

\subsection{Distilling Teacher Guidance}
\label{supp:lfa_guidance}

Following guidance distillation~\cite{meng2023distillation},
LFA uses the guided endpoint of the frozen SPA teacher as
an auxiliary supervision target.
Since paired SAR-to-EO translation prioritizes fidelity to
the observed scene, we use a fixed teacher guidance schedule.
We chose a CFG scale of $2.5$ as a moderate default,
independently of test-set performance, and apply it only
at the two middle Euler steps, yielding
$\mathbf s=(1,2.5,2.5,1)$.
The four-step teacher thus uses six NFEs, while the one-
and two-step students require only one and two NFEs,
respectively, without inference-time CFG.

\begin{table*}[!t]
\centering
\GeoSuppTableStyle
\setlength{\tabcolsep}{3pt}

\caption{\textbf{Quality and inference speed after larger-patch
adaptation on QXS-SAROPT.}
P16 references are the SPA and LFA models from
Table~\mainref{tab:main_results}.
Latency is measured at $256\times256$ with batch size one.
All speedups are relative to P16 SPA 4-step.
At the same sampling budgets, the P32-adapted models are
approximately $1.36\times$ faster than the P16 references,
with lower quality on all five metrics.}
\label{tab:supp_patch_quality}

\begin{tabularx}{\textwidth}
{>{\raggedright\arraybackslash}Xccrrrrrrc}
\toprule
Model
& \shortstack{Total\\updates}
& \shortstack{Steps\\/ NFE}
& FID$\downarrow$
& DISTS$\downarrow$
& LPIPS$\downarrow$
& SSIM$\uparrow$
& PSNR$\uparrow$
& \shortstack{Latency\\(ms)$\downarrow$}
& \shortstack{Speedup\\$\uparrow$}\\
\midrule

P16 SPA 4-step
& 120k & 4 / 6
& \textbf{16.8} & .221 & .509 & .362 & 17.00
& 83.08 & $1.00\times$\\

P16 LFA 1-step
& 130k & 1 / 1
& 17.4 & \textbf{.213} & \textbf{.498}
& \textbf{.365} & \textbf{17.12}
& 14.00 & $5.94\times$\\

\midrule

\GeoSuppOurRow
P16 $\to$ P32 adaptation 4-step
& 180k & 4 / 6
& 28.5 & .238 & .536 & .341 & 16.42
& 60.98 & $1.36\times$\\

\GeoSuppOurRow
P16 $\to$ P32 adaptation 1-step
& 190k & 1 / 1
& 26.1 & .232 & .519 & .335 & 16.33
& \textbf{10.29} & $\mathbf{8.08\times}$\\

\bottomrule
\end{tabularx}
\end{table*}

\begin{figure*}[!t]
\centering
\includegraphics[width=\textwidth]
{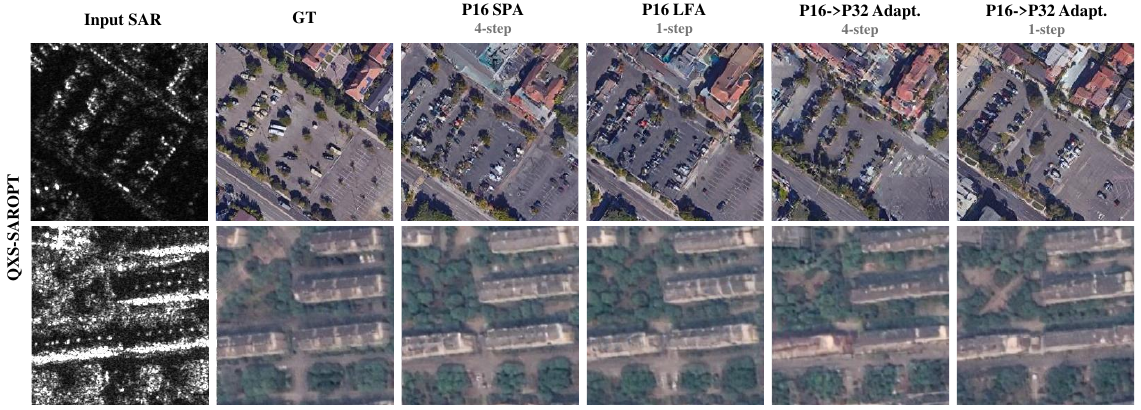}

\caption{\textbf{Qualitative comparison of P16 and P32-adapted
models on QXS-SAROPT.}
Columns show input SAR, GT EO, P16 SPA 4-step,
P16 LFA 1-step, P16 $\to$ P32 adaptation 4-step,
and P16 $\to$ P32 adaptation 1-step.
P16 models produce more plausible fine details in the
parking area and rooftop patterns, while P32-adapted
outputs retain the overall layout with coarser local detail.}
\label{fig:supp_patch_qualitative}
\vspace{-7px}
\end{figure*}

\subsection{Larger-Patch Adaptation}
\label{supp:patch_adaptation}

We investigate larger-patch adaptation as a potential route to
faster SAR-to-EO generation. P16 and P32 denote patch sizes of
$16\times16$ and $32\times32$ pixels, respectively.
Motivated by the comparable generation quality and faster
inference reported by Jiang et al.~\cite{jiang2026empirical},
we adapt both EO and SAR backbone streams from P16 to P32
while retaining the P16 PiT decoder.
This reduces the number of backbone tokens per stream to
one quarter at a fixed image resolution.

To transfer the pretrained P16 weights, we initialize the
larger input kernel of each stream
$m\in\{\mathrm{EO},\mathrm{SAR}\}$ by spatial tiling:
\begin{equation}
\mathbf W_{32}^{m}=\frac{1}{2}
\begin{bmatrix}
\mathbf W_{16}^{m} & \mathbf W_{16}^{m}\\
\mathbf W_{16}^{m} & \mathbf W_{16}^{m}
\end{bmatrix}.
\label{eq:supp_p32_kernel}
\end{equation}
The stride becomes 32, and all other weights are copied.
We connect the coarser backbone grid to the existing PiT
decoder without introducing additional parameters.
Let $\mathbf H^{(32)}$ denote the final shared-block EO
feature grid. Each P32 feature provides the PiT condition
at four P16 locations:
\begin{equation}
\mathbf C^{(16)}_{2i+a,\,2j+b}=\mathbf H^{(32)}_{ij},
\qquad a,b\in\{0,1\},
\label{eq:supp_p32_condition}
\end{equation}
where $(i,j)$ index the P32 grid from zero.
Backbone RoPE uses
$(2i+\tfrac12,2j+\tfrac12)$ in P16-grid units,
while PiT retains its P16 positions.

To align REPA supervision with the coarser token grid,
we average-pool the frozen DINO features
$\mathbf Z^{(16)}=\mathcal F_\phi^{\mathrm{last}}(\mathbf x^e)$
over each $2\times2$ neighborhood:
\begin{equation}
\bar{\mathbf Z}^{(32)}_{ij}
=\frac14
\sum_{a,b\in\{0,1\}}
\mathbf Z^{(16)}_{2i+a,\,2j+b}.
\label{eq:supp_p32_repa_pool}
\end{equation}
The pooled features replace the target in Eq.~\mainref{eq:repa}
for the P32 EO features $\mathbf h_e^L$ from the final
dual-stream block.

We initialize P16-to-P32 adaptation from the raw P16 SPA
checkpoint at 120k updates and jointly fine-tune both
backbone streams for 60k additional updates.
The resulting 180k raw checkpoint initializes the one-step
adaptation model, which is fine-tuned for another 10k updates.
Both P32 adaptation stages omit CPD and teacher-endpoint losses.
The P16 SPA and LFA references are the corresponding
models reported in Table~\mainref{tab:main_results}.

Table~\ref{tab:supp_patch_quality} compares the P32-adapted
models with these P16 references.
At the four-step sampling budget, latency decreases from
$83.08$ to $60.98$\,ms, a $1.36\times$ speedup.
For one-step generation, latency decreases from
$14.00$ to $10.29$\,ms, also yielding a $1.36\times$ speedup.
These correspond to latency reductions of $26.6\%$ and
$26.5\%$, respectively.
Relative to P16 SPA 4-step, P16 LFA 1-step and
P16-to-P32 adaptation 1-step are approximately
$5.9\times$ and $8.1\times$ faster.
The latter gain combines larger-patch adaptation with
the reduction in sampling steps.

The tested P32-adapted models obtain worse values on all
five quality metrics than the corresponding P16 references.
For four-step generation, FID increases from $16.8$ to
$28.5$, DISTS from $.221$ to $.238$, and PSNR decreases
from $17.00$ to $16.42$\,dB.
Within the P32 setting, dedicated one-step adaptation
improves FID, DISTS, and LPIPS relative to the four-step
model, while slightly reducing SSIM and PSNR.
The resulting one-step model still underperforms P16 LFA
1-step on every quality metric: for example, FID is
$26.1$ versus $17.4$, DISTS is $.232$ versus $.213$,
and PSNR is $16.33$ versus $17.12$\,dB.

Figure~\ref{fig:supp_patch_qualitative} illustrates the
corresponding differences in local detail.
In the shown examples, P16 models produce more plausible
fine structures, including small objects in the parking
area in the first row and rooftop patterns in the second.
The P32-adapted models retain the broad scene layout,
but some local structures and textures appear coarser
or less faithful to the reference EO.
These observations are consistent with the quantitative
quality differences.
The tested P32 variants offer lower latency, while the
P16 references preserve better structural and perceptual
fidelity, supporting our use of P16 for the main models.

\section{Interpreting Evaluation Metrics and Qualitative Results}
\label{supp:qualitative}

\begin{figure*}[!t]
\centering
\includegraphics[width=\textwidth]{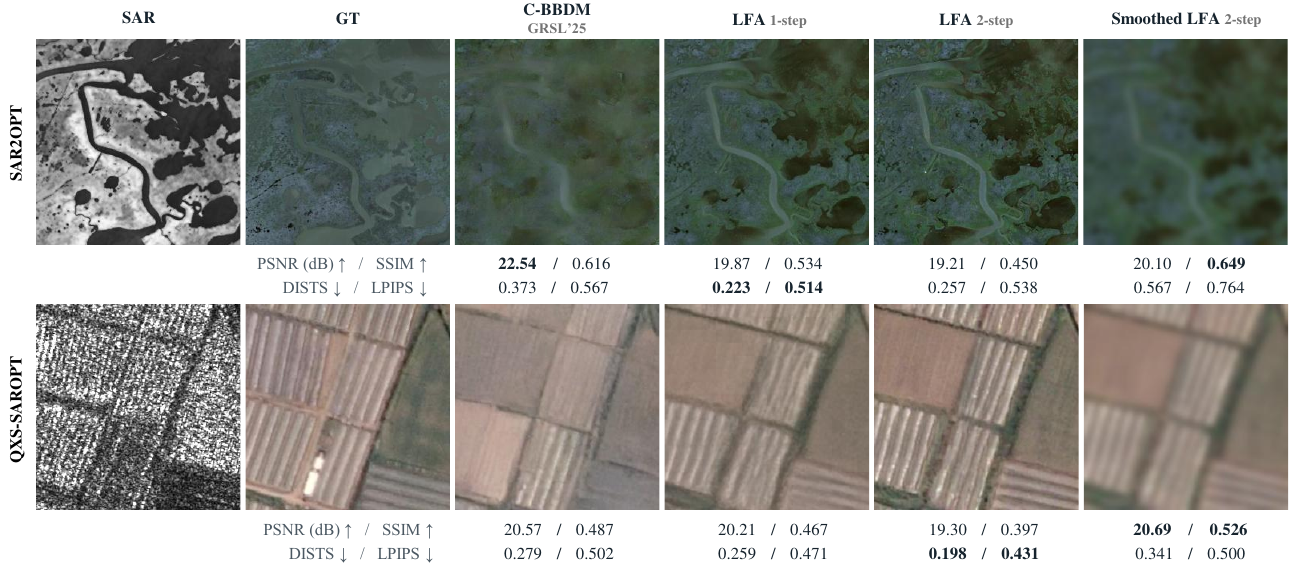}
\caption{\textbf{Effect of smoothing on SAR-to-EO translation metrics.}
The top and bottom rows show SAR2Opt and QXS-SAROPT, respectively.
The last two columns compare the same two-step LFA prediction
before and after smoothing. Values below each prediction are
per-image scores against the paired EO reference.}
\label{fig:supp_metric_examples}
\end{figure*}

\subsection{Interpreting PSNR and SSIM in SAR-to-EO Translation}
\label{supp:metric_interpretation}

We smooth each two-step LFA prediction by averaging neighboring
pixel values within each color channel, then compare the original
and smoothed predictions against the same EO reference
(Fig.~\ref{fig:supp_metric_examples}). In both scenes, smoothing
raises PSNR and SSIM while increasing DISTS and LPIPS. In the
QXS-SAROPT scene, for example, the parallel field strips become
harder to distinguish as SSIM rises from 0.397 to 0.526 and DISTS
worsens from 0.198 to 0.341.

In both scenes, the cBBDM output and the smoothed two-step LFA
prediction each achieve higher PSNR and SSIM than either of the
two unsmoothed LFA predictions. The unsmoothed one- and two-step
LFA predictions, however, retain finer detail and achieve lower
DISTS and LPIPS. We therefore prioritize DISTS and LPIPS when
assessing perceptual similarity to the paired EO reference and
use FID to assess distributional quality across the test set.
When methods perform similarly on these measures, we use PSNR
and SSIM to compare reference fidelity.

\subsection{Comparison across EPT, SPA, and LFA}
\label{supp:qualitative_phases}

Figure~\ref{fig:supp_stage_qualitative} compares four-step
EPT and SPA with independently trained one- and two-step
LFA models.
In the shown QXS-SAROPT comparison, the full SPA model
follows SAR spatial cues more closely than EPT in regions
where the SAR observation and reference EO differ.
Both LFA models retain comparable structural fidelity
and visual quality with fewer sampling steps.

The SAR2Opt example also illustrates a limitation.
EPT achieves the lowest FID and DISTS among these four
models (Table~\mainref{tab:main_results}) and more faithfully
reproduces the scene semantics in the shown example.
Following replacement of the SAR VAE interface, SPA
introduces semantic errors despite CPD supervision:
part of the sparsely vegetated ground is rendered as
denser vegetation.
Related errors persist to varying degrees in both LFA
models.
This example shows that CPD does not fully prevent
semantic degradation during SAR pixel conversion,
and subsequent low-step adaptation can inherit
these limitations.

\subsection{Additional Qualitative Comparison with Baselines}
\label{supp:qualitative_baselines}

Figures~\ref{fig:supp_qxs_qual1} and~\ref{fig:supp_qxs_qual2}
present two additional QXS-SAROPT examples, while
Figs.~\ref{fig:supp_sar2opt_qual1} and~\ref{fig:supp_sar2opt_qual2}
present two SAR2Opt examples.
Each figure shows a single scene, including the SAR input,
paired EO reference, outputs of selected baselines from
Table~\mainref{tab:main_results}, and our LFA 1-step and 2-step results.
These examples provide additional comparisons of scene
structure, object boundaries, and local texture.

\begin{figure*}[!t]
\centering
\includegraphics[width=0.965\textwidth]
{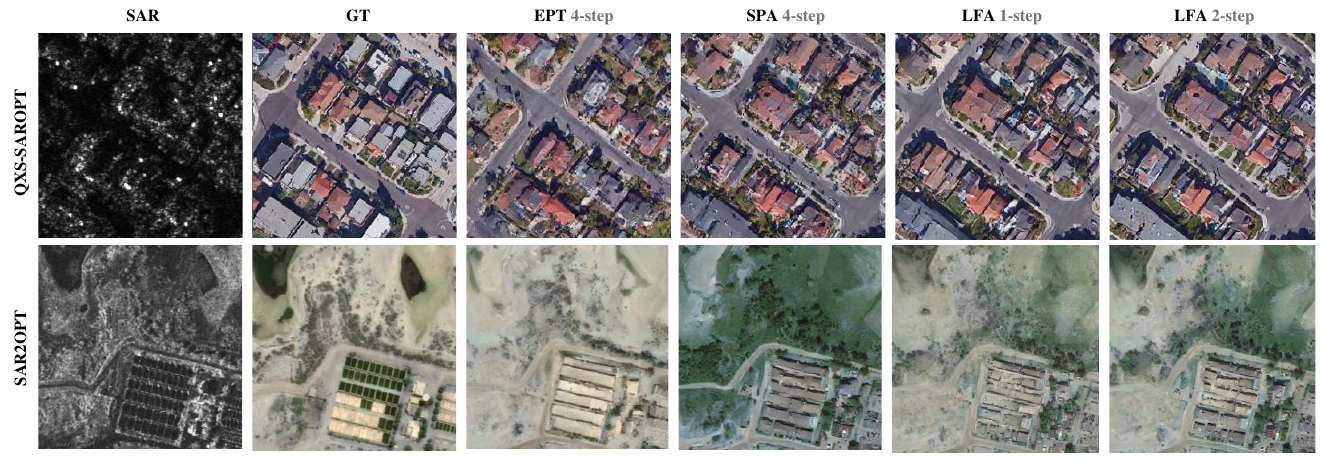}
\caption{\textbf{Qualitative comparison across EPT, SPA, and LFA.}
Top: QXS-SAROPT, where SPA follows SAR spatial cues more
closely and LFA retains comparable quality.
Bottom: a SAR2Opt failure case, where semantic errors
appear during SAR pixel conversion despite CPD and
partly persist in both LFA models.
EPT and SPA use four steps; LFA uses one or two.}
\label{fig:supp_stage_qualitative}
\end{figure*}

\begin{figure*}[!t]
\centering
\includegraphics[width=0.965\textwidth]
{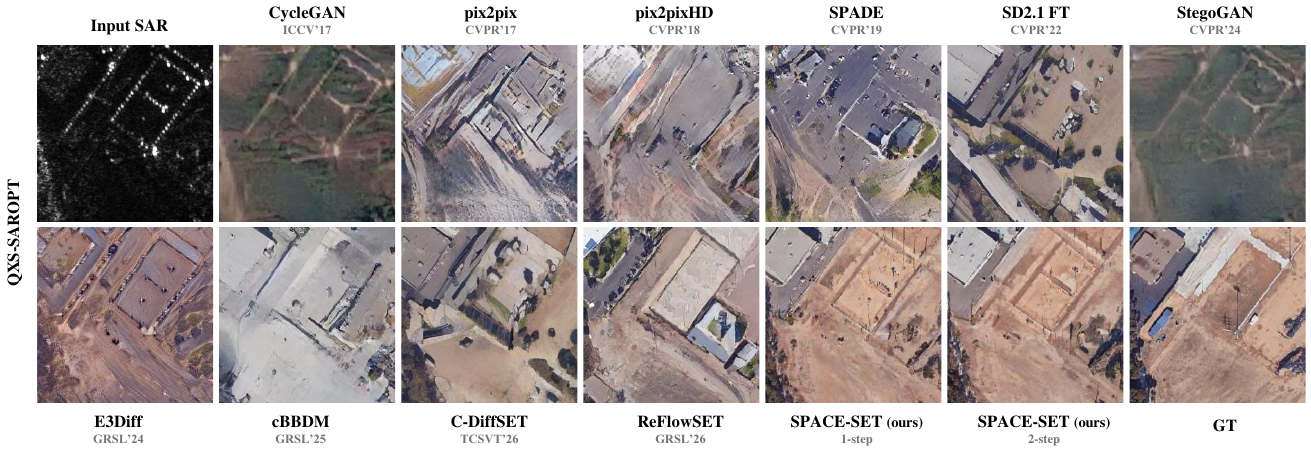}
\caption{\textbf{Additional qualitative comparison on
QXS-SAROPT (example 1).}
The SAR input and paired EO reference are shown alongside
baseline outputs and our LFA 1-step and 2-step results.}
\label{fig:supp_qxs_qual1}
\end{figure*}

\begin{figure*}[!t]
\centering
\includegraphics[width=0.965\textwidth]
{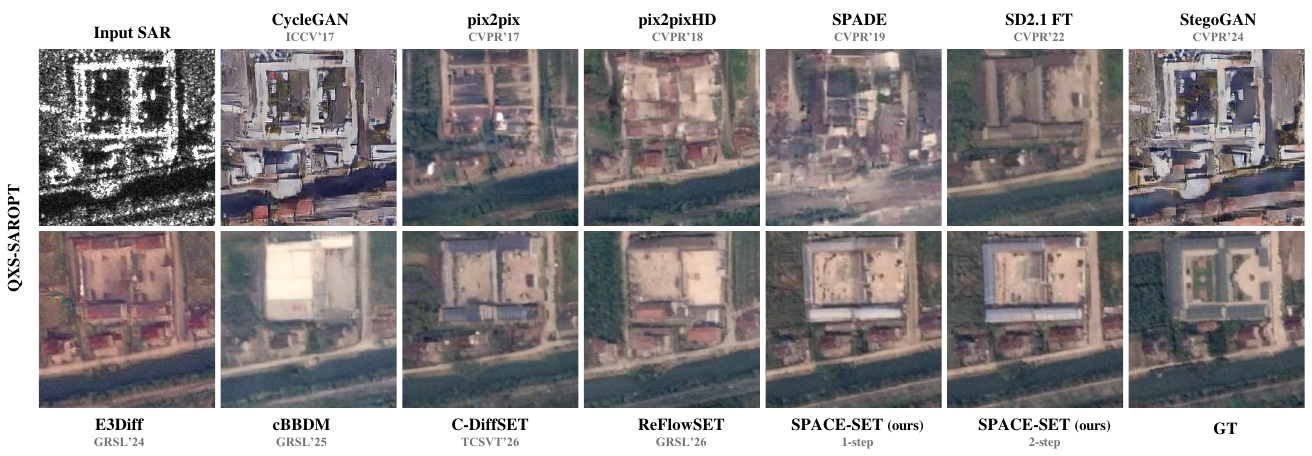}
\caption{\textbf{Additional qualitative comparison on
QXS-SAROPT (example 2).}
The SAR input and paired EO reference are shown alongside
baseline outputs and our LFA 1-step and 2-step results.}
\label{fig:supp_qxs_qual2}
\end{figure*}

\begin{figure*}[!t]
\centering
\includegraphics[width=0.965\textwidth]
{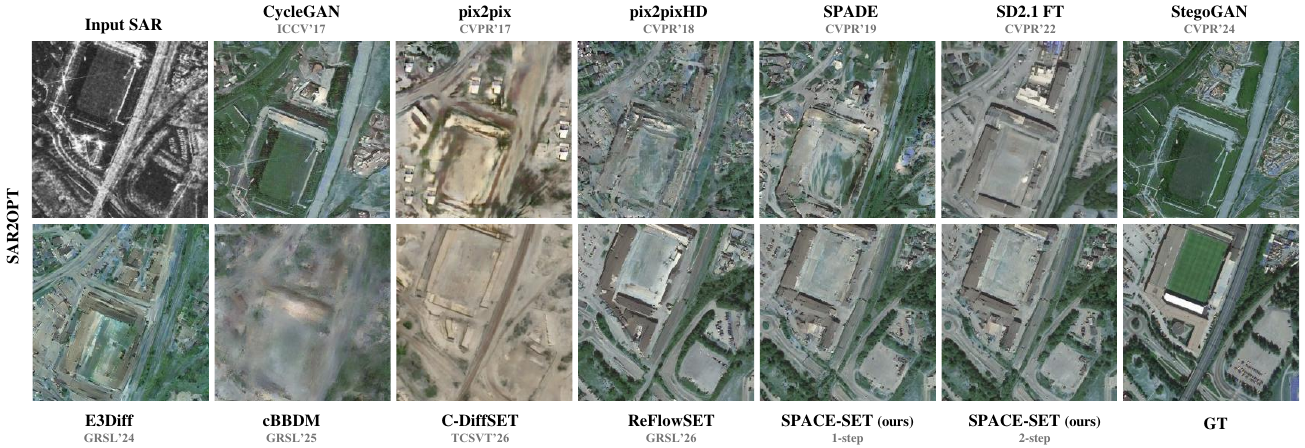}
\caption{\textbf{Additional qualitative comparison on
SAR2Opt (example 1).}
The SAR input and paired EO reference are shown alongside
baseline outputs and our LFA 1-step and 2-step results.}
\label{fig:supp_sar2opt_qual1}
\end{figure*}

\begin{figure*}[!t]
\centering
\includegraphics[width=0.965\textwidth]
{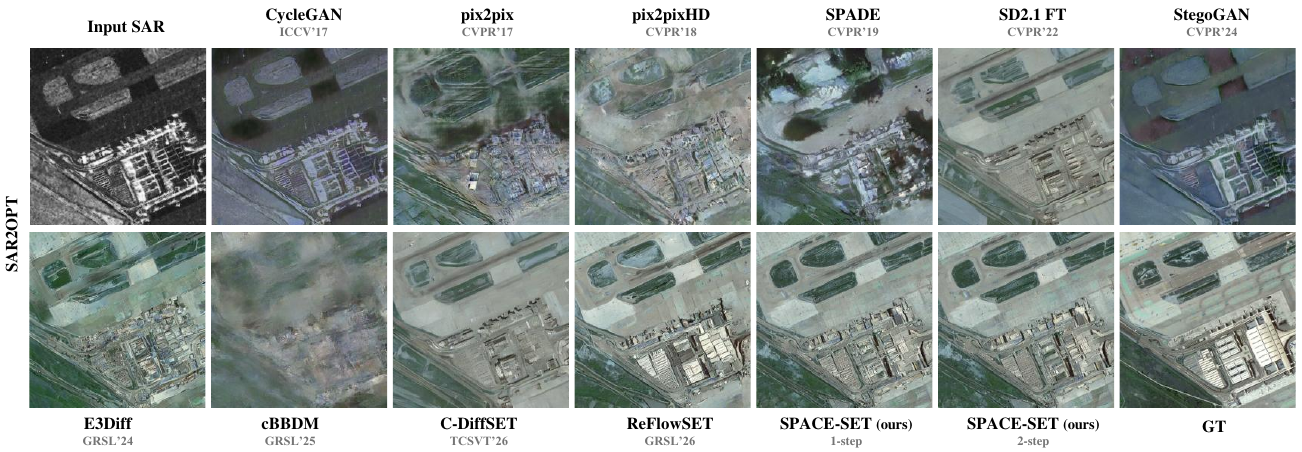}
\caption{\textbf{Additional qualitative comparison on
SAR2Opt (example 2).}
The SAR input and paired EO reference are shown alongside
baseline outputs and our LFA 1-step and 2-step results.}
\label{fig:supp_sar2opt_qual2}
\end{figure*}

\section{Limitations and Future Work}
\label{supp:limitations}

\paragraph{SAR-specific semantic priors.}
Our SAR pathway inherits features from the pretrained
SAR-to-EO generator. The SAR detail-injection branch
offers a way to test whether richer conditioning helps.
In place of the lightweight CNN, a SAR-pretrained encoder
such as SARMAE~\cite{liu2026sarmae} could supply spatial features.
SAR--EO, radar--optical, and SAR--language pretraining offer
other sources of supervision~\cite{do2026sareofm,fuller2023croma,jiang2026sarclip}.
Running a separate encoder would add a forward pass per
SAR image, even if its features were reused at every sampling
step. Comparing translation quality and end-to-end latency at
one and two steps would establish whether this cost is
justified. Training-time distillation into the existing SAR
pathway could transfer these features while preserving the
deployed architecture.

\paragraph{Training complexity.}
The sequential EPT, SPA, and LFA phases, together with
the supervision curriculum, require multiple training
transitions and phase-specific settings.
Future work could explore a unified training procedure
that jointly learns pixel-space conversion and low-step
generation.
Adaptive objective scheduling may reduce manual phase
transitions while preserving the benefits of gradual
representation transfer.

\paragraph{Student architecture.}
LFA reduces the number of sampling steps while retaining
the SPA generator architecture.
A complementary way to accelerate inference is to reduce
the cost of each forward pass by distilling the SPA teacher
into a lightweight student with a compact backbone and
pixel decoder.
The student capacity and distillation objectives could be
jointly optimized to preserve structural and perceptual fidelity.
Such a faster one-step model could be of strategic value
in time-sensitive aerospace operations, for example by
providing low-latency EO-like views of airborne or satellite
SAR imagery for rapid disaster assessment.

\end{document}